\documentclass[11pt,a4paper, dvipsnames]{article}
\usepackage[hyperref]{emnlp2020}
\usepackage{times}
\usepackage{latexsym}

\usepackage{diagbox}
\usepackage{enumitem}
\usepackage{textcomp}
\usepackage[export]{adjustbox}
\usepackage{tikz}
\usepackage{rotating}
\usetikzlibrary{positioning, arrows, calc}
\usepackage{amsmath}
\usepackage{booktabs}
\usepackage{amssymb}
\usepackage{multirow}
\usepackage{graphicx}
\usepackage{url}
\usepackage{tabularx}
\usepackage{listings}
\usepackage{microtype}

\aclfinalcopy 

\title{E-BERT: Efficient-Yet-Effective Entity Embeddings for BERT}

\author{Nina Poerner$^{\ast\dagger}$ and Ulli Waltinger$^\dagger$ and Hinrich Sch{\"u}tze$^\ast$ \\
  $^\ast$Center for Information and Language Processing, LMU Munich, Germany \\
  $^\dagger$Corporate Technology Machine Intelligence (MIC-DE), Siemens AG Munich, Germany \\
{\tt poerner@cis.uni-muenchen.de | inquiries@cislmu.org} }

\date{}

\def\figref#1{Figure~\ref{fig:#1}}
\def\figlabel#1{\label{fig:#1}\label{p:#1}}

\def\tabref#1{Table~\ref{tab:#1}}
\def\tablabel#1{\label{tab:#1}\label{p:#1}}

\def\secref#1{Section~\ref{sec:#1}}
\def\seclabel#1{\label{sec:#1}}
\def\eqref#1{Eq.~\ref{eqn:#1}}

\def\eqlabel#1{\label{eqn:#1}}

\begin{document}
\maketitle
\begin{abstract}
We present a novel way of injecting factual knowledge about entities into the pretrained BERT model \cite{devlin2019bert}:
We align Wikipedia2Vec entity vectors \cite{yamada2016joint}
with BERT's native wordpiece vector space and use the aligned entity vectors as if they were wordpiece vectors.
The resulting entity-enhanced version of BERT (called \textbf{E-BERT}) is similar in spirit to ERNIE \cite{zhang2019ernie} and KnowBert \cite{peters2019knowledge}, but it requires no expensive further pretraining of the BERT encoder.
We evaluate E-BERT on
unsupervised question answering (QA),
supervised
relation
classification (RC) and entity linking (EL).
On all three tasks,
E-BERT outperforms BERT and other baselines.
We also show quantitatively that the original BERT model is overly reliant on the surface form of entity names (e.g.,
guessing that someone with an Italian-sounding name speaks Italian), 
and that E-BERT mitigates this problem.
\end{abstract}

\section{Introduction}
\seclabel{intro} BERT \cite{devlin2019bert} and its
successors (e.g., \citet{yang2019xlnet, liu2019roberta,
  wang2019structbert}) continue to achieve state of the art
performance on various NLP tasks.  Recently, there has
been interest in enhancing BERT with factual knowledge about entities
\cite{zhang2019ernie, peters2019knowledge}.  To this end, we
introduce \textbf{E-BERT}: 
We align Wikipedia2Vec entity vectors \cite{yamada2016joint}
with BERT's wordpiece vector space (\secref{wikipedia2vec}) and feed the aligned vectors into BERT as if they were wordpiece vectors (\secref{injection}).
Importantly, we do not make any changes to the BERT encoder itself, and we do no 
additional pretraining.
This stands in contrast to previous entity-enhanced versions of BERT,
such as ERNIE or KnowBert, which require additional encoder pretraining.

In \secref{qa}, we evaluate our approach on
LAMA \cite{petroni2019language},
a recent
unsupervised QA benchmark for pretrained Language Models (LMs). We set a new state of the art on LAMA, with improvements over original BERT, ERNIE and KnowBert.
We also find
that the original BERT model is overly reliant
on the surface form of
entity names, e.g., it predicts that a person with
an Italian-sounding name speaks Italian, regardless of whether this is factually correct.
To quantify this effect, 
we create \textbf{LAMA-UHN} (UnHelpfulNames), a
subset of LAMA where questions with overly helpful entity names were deleted
(\secref{filtering_lama}).

In \secref{downstream}, we show how to apply E-BERT to two entity-centric downstream tasks: relation classification (\secref{re}) and entity linking (\secref{el}).
On the former task, we feed aligned entity vectors as inputs, on the latter, they serve as inputs \textit{and} outputs.
In both cases, E-BERT outperforms original BERT and other baselines.

\paragraph{Summary of contributions.}
\begin{itemize}
 \setlength\itemsep{0mm}
\item Introduction of E-BERT: Feeding entity vectors
  into BERT without additional encoder pretraining. (\secref{method})
\item Evaluation on the LAMA unsupervised QA benchmark: E-BERT
  outperforms  BERT, ERNIE and KnowBert. (\secref{qa})
\item LAMA-UHN: A harder version of the LAMA benchmark with less informative entity names.
(\secref{filtering_lama})
\item Evaluation on supervised relation classification (\secref{re}) and entity
  linking (\secref{el}).
\item Upon publication, we will release LAMA-UHN as well as E-BERT$_\mathtt{BASE}$ and E-BERT$_\mathtt{LARGE}$.\footnote{\url{https://github.com/anonymized}}
\end{itemize}

\section{Related work}
\subsection{BERT}
BERT (Bidirectional Encoder Representations from Transformers) is a Transformer  \cite{vaswani2017attention} that was pretrained as a masked LM (MLM) on unlabeled text.
At its base, BERT segments text into wordpieces from a vocabulary 
$\mathbb{L}_\mathrm{WP}$.
Wordpieces are embedded into real-valued vectors by a lookup
function (denoted $\mathcal{E}_\mathrm{BERT} :
\mathbb{L}_\mathrm{WP}
\rightarrow \mathbb{R}^{d_\mathrm{BERT}}$).
The wordpiece vectors are combined with position and segment embeddings and then fed into a stack of Transformer layers (the encoder, denoted $\mathcal{F}_\mathrm{BERT}$).
During pretraining, some wordpieces are replaced by a special \textit{[MASK]} token.
The output of BERT is fed into a final feed-forward net (the MLM head, denoted $\mathcal{F}_\mathrm{MLM}$), to predict the identity of the masked wordpieces.
After pretraining, the MLM head is usually replaced by a task-specific layer, and the entire model is finetuned on supervised data.

\subsection{Entity-enhanced BERT}
This paper adds to recent work on entity-enhanced BERT models, most notably ERNIE \cite{zhang2019ernie} and KnowBert \cite{peters2019knowledge}.
ERNIE and KnowBert are based on the design principle that \textit{BERT be adapted to entity vectors}: 
They introduce new encoder layers to feed pretrained entity vectors into the Transformer, and they require additional pretraining to integrate the new parameters.
In contrast, E-BERT's design principle is that \textit{entity vectors be adapted to BERT}, which makes our approach more efficient (see \secref{systems}).

Two other knowledge-enhanced MLMs are KEPLER \cite{wang2019kepler} and K-Adapter \cite{wang2020k}, which are based on Roberta \cite{liu2019roberta} rather than BERT.
Their factual knowledge does not stem from entity vectors -- instead, they are trained in a multi-task setting on relation classification and knowledge base completion.

\subsection{Wikipedia2Vec}
\seclabel{wiki2vec} 
Wikipedia2Vec \cite{yamada2016joint} embeds words and
entities (Wikipedia URLs) into a common space.
Given a vocabulary of words
$\mathbb{L}_\mathrm{Word}$ and a vocabulary of entities
$\mathbb{L}_\mathrm{Ent}$, it learns a lookup embedding function
$\mathcal{E}_\mathrm{Wikipedia}: \mathbb{L}_\mathrm{Word}
\cup \mathbb{L}_\mathrm{Ent} \rightarrow
\mathbb{R}^{d_\mathrm{Wikipedia}}$.
The Wikipedia2Vec loss has three components:
(1) skipgram Word2Vec \cite{mikolov2013efficient} operating on $\mathbb{L}_\mathrm{Word}$, (2) a graph loss operating on the Wikipedia hyperlink graph, whose vertices are $\mathbb{L}_\mathrm{Ent}$ and (3) a version of Word2Vec where words are predicted from entities.
Loss (3) ensures that entities and words are embedded into the same space.

\subsection{Vector space alignment}
Our vector space alignment strategy is inspired by cross-lingual word vector alignment (e.g., \citet{mikolov2013exploiting, smith2017offline}).
A related method was recently applied by \citet{wang2019improving} to map cross-lingual word vectors into the multilingual BERT wordpiece vector space.

\subsection{Unsupervised QA}
QA has typically been tackled as a supervised problem
(e.g., \citet{das2017question, sun2018open}).  
Recently, there has been interest in using unsupervised LMs such as GPT-2 or BERT for this task \cite{radford2019language, petroni2019language}.
\citet{davison2019commonsense} mine unsupervised commonsense knowledge from BERT, and \citet{jiang2019how} show the importance of using good prompts for unsupervised QA.
None of this prior work differentiates quantitatively
between factual knowledge
of LMs and their ability to reason about the
surface form of entity names.

\section{E-BERT}
\seclabel{method}
\subsection{Aligning entity and wordpiece vectors}
\seclabel{wikipedia2vec}
Conceptually, we want to transform the vectors of the entity vector space $\mathcal{E}_\mathrm{Wikipedia}[\mathbb{L}_\mathrm{Ent}]$ in such a way that they look to BERT like vectors from its native wordpiece vector space $\mathcal{E}_\mathrm{BERT}[\mathbb{L}_\mathrm{WP}]$.
We model the transformation as an unconstrained linear mapping $\mathbf{W} \in \mathbb{R}^{d_\mathrm{BERT} \times d_\mathrm{Wikipedia}}$.
Since $\mathbb{L}_\mathrm{WP}$ does not contain any entities (i.e., $\mathbb{L}_\mathrm{WP} \cap \mathbb{L}_\mathrm{Ent} = \{\}$), we fit the mapping on $\mathbb{L}_\mathrm{WP} \cap \mathbb{L}_\mathrm{Word}$:
$$
\sum_{x \in \mathbb{L}_\mathrm{WP} \cap \mathbb{L}_\mathrm{Word}} ||\mathbf{W}\mathcal{E}_\mathrm{Wikipedia}(x)-\mathcal{E}_\mathrm{BERT}(x)||_2^2
$$
Since Wikipedia2Vec embeds $\mathbb{L}_\mathrm{Word}$ and $\mathbb{L}_\mathrm{Ent}$ into the same space (see \secref{wiki2vec}), $\mathbf{W}$ can be applied to $\mathbb{L}_\mathrm{Ent}$ as well.
We define the E-BERT embedding function as:
\begin{align}
	\nonumber
	& \mathcal{E}_\mathrm{E\text{-}BERT}: \mathbb{L}_\mathrm{Ent} \rightarrow \mathbb{R}^{d_\mathrm{BERT}} \\
	\nonumber
	& \mathcal{E}_\mathrm{E\text{-}BERT}(a) = \mathbf{W} \mathcal{E}_\mathrm{Wikipedia}(a)
\end{align}

\subsection{Using aligned entity vectors}
\seclabel{injection}
We explore two strategies for feeding the aligned entity vectors into the BERT encoder:

\paragraph{E-BERT-concat.} 
E-BERT-concat combines entity IDs and wordpieces
by string concatenation, with the slash symbol as separator \cite{schick2019bertram}.
For example, the wordpiece-tokenized input
\begin{itemize}[leftmargin=*]
\small \item[] \textit{The native language of Jean Mara
  \#\#is is [MASK] .}\footnote{For readability, we omit the special tokens \textit{[CLS]} and \textit{[SEP]} from all examples.}
\end{itemize}
becomes
\begin{itemize}[leftmargin=*]
\small \item[] \textit{The native language of} \textbf{Jean\_Marais} \textit{/ Jean Mara \#\#is is [MASK] .}
\end{itemize}
The entity ID (\textbf{bold}) is embedded by $\mathcal{E}_\mathrm{E\text{-}BERT}$ and all wordpieces (\textit{italics}) are embedded by $\mathcal{E}_\mathrm{BERT}$ (see \figref{schematic}).
After the embedding operation, the sequence of vectors is combined with position and segment embeddings and fed into $\mathcal{F}_\mathrm{BERT}$, just like any normal sequence of wordpiece vectors.

E-BERT-concat is comparable to ERNIE or KnowBert, which also represent entities as a combination of surface form (wordpieces) and entity vectors.
But in contrast to ERNIE and KnowBERT, \textbf{we do not change or further pretrain the BERT encoder itself}.

\paragraph{E-BERT-replace.}
For ablation purposes, we define another variant of E-BERT that substitutes the entity surface form with the entity vector.
With E-BERT-replace, our example becomes:
\begin{itemize}[leftmargin=*]
\small \item[] \textit{The native language of} \textbf{Jean\_Marais} \textit{is [MASK] .}
\end{itemize}

\subsection{Implementation}
\seclabel{systems} We train cased Wikipedia2Vec on a recent
Wikipedia dump (2019-09-02), setting $d_\mathrm{Wikipedia} =
d_\mathrm{BERT}$.  We ignore Wikipedia
pages with fewer than $5$ links (Wikipedia2Vec's default), with the exception
of entities needed for the downstream entity linking
experiments (see \secref{el}).
This results in an entity vocabulary of size $|\mathbb{L}_\mathrm{Ent}| = 2.7$M.\footnote{Due to the link threshold and some Wikidata-Wikipedia mismatches, we lack entity vectors for $6\%$ of LAMA questions and $10\%$ of FewRel sentences (RC experiment, see \secref{re}).
In these cases, we fall back onto using wordpieces only, i.e., onto standard BERT behavior.}

\paragraph{Computational cost.}
Training Wikipedia2Vec took us $\sim$6 hours on 32 CPUs, and the cost of fitting the linear transformation $\mathbf{W}$ is negligible.
We did not require a GPU.
For comparison, KnowBert W+W was pretrained for 1.25M steps on up to four Titan RTX GPUs, and ERNIE took one epoch on the English Wikipedia. (ERNIE's pretraining hardware was not disclosed, but it seems likely that a GPU was involved.)

\begin{figure}
\centering
\begin{tikzpicture}[
  aligned/.style={%
    text height=.75ex,
    text depth=.25ex,
    text centered
  }
]
\node [aligned, ] at (0.1,-.85) (w1) {\tiny \textit{The}};
\node [aligned, right=.3mm of w1] (w2) {\tiny \textit{native}};
\node [aligned, right=.3mm of w2] (w3) {\tiny \textit{language}};
\node [aligned, right=.3mm of w3] (w4) {\tiny \textit{of}};
\node [aligned, right=.3mm of w4] (w5) {\tiny \textbf{Jean\_Marais}};
\node [aligned, right=.3mm of w5] (w6) {\tiny \textit{/}};
\node [aligned, right=1.5mm of w6] (w7) {\tiny \textit{Jean}};
\node [aligned, right=.3mm of w7] (w8) {\tiny \textit{Mara}};
\node [aligned, right=.3mm of w8] (w9) {\tiny \textit{\#\#is}};
\node [aligned, right=.1mm of w9] (w10) {\tiny \textit{...}};

\node at (.8, -3) [rectangle, draw, fill = blue!50] (lm) {\small $\mathcal{E}_\mathrm{BERT}[\mathbb{L}_\mathrm{WP}]$};
\foreach \i in {1,...,4}{
\draw [->, >=stealth, thick, dashed, gray, bend angle=20, bend left] ([xshift=-20pt] lm.north) to (w\i);}
\foreach \i in {6,...,9}{
\draw [->, >=stealth, thick, dashed, gray, bend angle=20, bend right] (lm.east) to ([yshift=2pt] w\i.south);}

\foreach \i in {1,2,3,4,6,7,8,9}{
\node [above=-.3mm of w\i, rectangle, minimum height=3mm, draw, fill=blue!50] (e\i) {};
}
\node [above=-.3mm of w5, rectangle, minimum height=3mm, draw, fill=red!50] (e5) {};

\node at (6.1, -3) [rectangle, draw, fill = gray!50] (e) {\small $\mathcal{E}_\mathrm{Wikipedia}[\mathbb{L}_\mathrm{Ent}]$};
\node (lower) at (1, -.1) {};
\node (upper) at (1, .4) {};

\draw (lm.west |- lower) [draw, fill=gray!50] rectangle  (e.east |- upper) node[pos=.5] {\small $\mathcal{F}_\mathrm{BERT}$ \tiny (BERT encoder)};
\node at (0,-.1) (ref) {};
\foreach \i in {1,...,9} {
\draw (e\i) [->, >=stealth, thick] -- (e\i|-ref);
}

\node at ($(e.west)!0.5!(lm.east)$) [rectangle, draw, fill = gray!50] (w) {\small $\mathcal{E}_\mathrm{Wikipedia}[\mathbb{L}_\mathrm{Word}]$};
\node at (2.25, -1.5) [rectangle, draw, fill=gray!50] (trans) {\small $\mathbf{W}$};
\node at (1.75, -4) [rectangle, draw, fill = gray!50] (bert) {\small BERT wordpiece layer};
\node at (5, -4) [rectangle, draw, fill = gray!50] (wiki) {\small Wikipedia2Vec};
\node [align=center] at (2.15, -2.2) {\tiny (linear transformation \\[-2.4mm] \tiny from $\mathcal{E}_\mathrm{Wikipedia}$ to $\mathcal{E}_\mathrm{BERT}$  \\[-2.2mm] \tiny fitted on $\mathbb{L}_\mathrm{WP} \cap \mathbb{L}_\mathrm{Word}$)};
\node [align=center, rectangle, draw, fill=red!50, above=4.25mm of e.north east,anchor=south east] (transformed) {\small $\mathcal{E}_\mathrm{E\text{-}BERT}[\mathbb{L}_\mathrm{Ent}] = $ \\ \small $\mathbf{W} \mathcal{E}_\mathrm{Wikipedia}[\mathbb{L}_\mathrm{Ent}]$};
\draw [->, >=stealth, thick, bend angle=40, bend right] ([xshift=13pt] w.north) to ([yshift=-4pt] trans.east);
\draw [->, >=stealth, thick, bend angle=40, bend left] ([xshift=-12pt] lm.north) to ([yshift=-4pt] trans.west);
\node [below=0.3mm of lm, inner sep=0] (lmlabel) {\tiny (wordpiece vector space)};
\node [below=0.3mm of w, inner sep=0] (wlabel) {\tiny (word vector space)};
\node [below=0.3mm of e, inner sep=0] (elabel) {\tiny (entity vector space)};
\node [inner sep=0mm] (transformedtmp) at ([xshift=11pt] transformed.south) {};
\node [below=0.2mm of transformed.south east, anchor = north east, inner sep=0] (transformedlabel) {\tiny (aligned entity vector space)};
\draw (trans.east) [->, >=stealth, thick] -- (transformed.west|-trans.east);
\draw (e.north-|transformedlabel.south) [->, >=stealth, thick] -- (transformedlabel);
\draw [->, >=stealth, thick, bend angle=45, bend left] (bert.west) to ([xshift=-25pt] lmlabel.south);
\draw [->, >=stealth, thick, bend angle=45, bend left] (wiki.west) to ([xshift=7pt] wlabel.south);
\draw [->, >=stealth, thick, bend angle=45, bend right] (wiki.east) to ([xshift=10pt] elabel.south);
\draw [->, >=stealth, very thick, dashed, gray, bend angle=20, bend left] ([yshift=-3pt] transformed.north west) to ([yshift=2pt] w5.south);
\end{tikzpicture}
\caption{Schematic depiction of E-BERT-concat.}
\figlabel{schematic}
\end{figure}
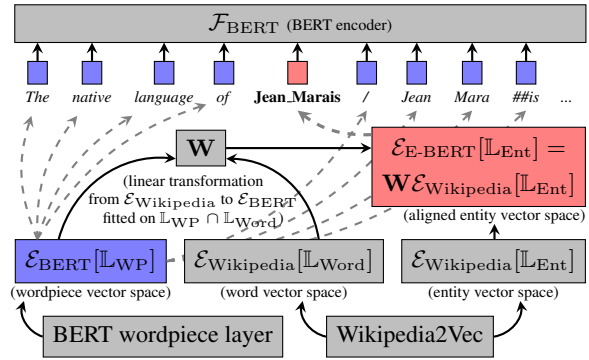

\section{Unsupervised QA}
\seclabel{qa}
\subsection{Data}
\seclabel{lama}
The LAMA (LAnguage Model Analysis) benchmark \cite{petroni2019language} probes for ``factual and commonsense knowledge'' of pretrained LMs.
In this paper, we use LAMA-Google-RE and LAMA-T-\mbox{REx} \cite{elsahar2018t}, which are aimed at factual knowledge.
Contrary to most previous work on QA, LAMA tests LMs without supervised finetuning.
\citet{petroni2019language} claim that BERT's performance on LAMA is comparable with a knowledge base (KB) automatically extracted from text, and speculate that BERT and similar models ``might become a viable alternative'' to such KBs.

The LAMA task follows this schema:
Given a KB triple (\textbf{sub}, \textbf{rel}, \textbf{obj}), the object is elicited with a relation-specific cloze-style question, e.g., (\textbf{Jean\_Marais},
\textbf{native-language}, \textbf{French}) becomes: ``The
native language of Jean Marais is
[MASK].''\footnote{LAMA provides oracle entity IDs, however, they are not used by the BERT baseline. For a fair evaluation, we ignore them too and instead use the Wikidata query API (\url{https://query.wikidata.org}) to infer entity IDs from surface forms. See Appendix for details.}
The model predicts a probability distribution over a limited vocabulary $\mathbb{L}_\mathrm{LAMA} \subset \mathbb{L}_\mathrm{WP}$ to replace [MASK], which is evaluated against the surface form of the object (here: \textit{French}).

\seclabel{experiments}
\subsection{Baselines}
Our primary baselines are cased BERT$_\mathtt{BASE}$ and BERT$_\mathtt{LARGE}$\footnote{\url{https://github.com/huggingface/transformers}} as evaluated in \citet{petroni2019language}.
We also test ERNIE \cite{zhang2019ernie}\footnote{\url{https://github.com/thunlp/ERNIE}} and KnowBert W+W \cite{peters2019knowledge},\footnote{\url{https://github.com/allenai/kb}} two entity-enhanced BERT$_\mathtt{BASE}$-type models.\footnote{ERNIE and KnowBert are uncased models.
We therefore lowercase all questions for them and restrict predictions to the intersection of their wordpiece vocabulary with lowercased $\mathbb{L}_\mathrm{LAMA}$. As a result, ERNIE and KnowBert select answers from $\sim$18K candidates (instead of $\sim$21K), which should work in their favor.
We verify that all lowercased answers appear in this vocabulary, i.e., ERNIE and KnowBert are in principle able to answer all questions correctly.}
E-BERT, ERNIE and KnowBert have entity vocabularies of size 2.7M, 5M and 470K, respectively.
As this might put KnowBert at a disadvantage, \tabref{lama} also reports performance on the subset of questions whose gold subject is known to KnowBert.

\subsection{Evaluation measure}
We use the same evaluation measure as \citet{petroni2019language}:
For a given $k$, we count a question as $1$ if the correct answer is among the top-$k$ predictions and as $0$ otherwise. 
\citet{petroni2019language} call this measure Precision@k (P@k).
Since this is not in line with the typical use of the term ``precision'' in information retrieval \cite[p. 161]{manning2008introduction},
we call the evaluation measure Hits@k.
Like \citet{petroni2019language}, we first average within relations and then across relations.

\begin{table}
\centering
\scriptsize
\setlength{\tabcolsep}{4pt}
\begin{tabularx}{.99\columnwidth}{X|ccccc}
\toprule
& original & E-BERT- & E-BERT- & ERNIE & Know-\\
& BERT & replace & concat & & Bert \\ 
\midrule
Jean Marais & French & French & French & french & french \\
Daniel Ceccaldi & Italian & French & French & french & italian\\
Orane Demazis & Albanian & French & French & french & french\\
Sylvia Lopez & Spanish & French & Spanish & spanish & spanish \\
Annick Alane & English & French & French & english & english \\
\bottomrule
\end{tabularx}
\caption{Native language (LAMA-T-REx:P103) of French-speaking actors according to different models. Model size is \texttt{BASE}.}
\tablabel{frenchpoets}
\end{table}

\begin{table*}
\scriptsize
\centering
\begin{tabularx}{.99\textwidth}{ll|l|l|Xr}
\toprule
\multicolumn{2}{l|}{Heuristic} & Relation  & \% deleted & Example of a deleted question \\
\midrule
\multirow{3}{*}{1} & \multirow{3}{*}{string match filter} 
&T-REx:P176 (manufacturer) & 81\% & Fiat Multipla is produced by [MASK:Fiat]. \\
& &T-REx:P138 (named after) & 75\% & Christmas Island is named after [MASK:Christmas]. \\
& &T-REx:P1001 (applies to jurisdiction) & 73\% & Australian Senate is a legal term in [MASK:Australia]. \\
\midrule
\multirow{3}{*}{2} & \multirow{3}{*}{person name filter}
& T-REx:P1412 (language used) & 63\% & Fulvio Tomizza used to communicate in [MASK:Italian]. & (1,1) \\
& &T-REx:P103 (native language) & 58\% & The native language of Tommy Nilsson is [MASK:Swedish]. & (-,1) \\
& &T-REx:P27 (nationality) & 56\% & Harumi Inoue is a [MASK:Japan] citizen.  & (1,-) \\
\bottomrule
\end{tabularx}
\caption{Statistics and examples of LAMA questions with helpful entity names, which were deleted from LAMA-UHN. We show the top-3 most strongly affected relations per heuristic. Numbers in brackets indicate which part(s) of the person name triggered the person name filter, e.g., (-,1) means that the correct answer was ranked first for the person's last name, but was not in the top-3 for their first name.}
\tablabel{filtered}
\end{table*}

\begin{table*}
	\scriptsize
	\centering
	\begin{tabularx}{.99\textwidth}{X|l|ccccc|cccc}
		\toprule
		& \multicolumn{1}{r|}{Model size} & \multicolumn{5}{c|}{\texttt{BASE}} & \multicolumn{4}{c}{\texttt{LARGE}} \\ \cmidrule{2-11}
		& \multicolumn{1}{r|}{\multirow{2}{*}{\backslashbox{Dataset}{Model}}} & original & E-BERT- & E-BERT- & ERNIE & Know- & original & E-BERT- & E-BERT- & K- \\
		& & BERT & replace & concat & & Bert & BERT & replace & concat & Adapter \\ \midrule
		
		\multirow{3}{*}{\parbox{8mm}{All\\questions}} & 0 (original LAMA) & 29.2 & 29.1 & \textbf{36.2} & 30.4 & 31.7 & 30.6 & 28.5 & 34.2 & 27.6 \\
		& 1 (string match filter) & 22.3 & 29.2 & \textbf{32.6} & 25.5 & 25.6 & 24.6 & 28.6 & 30.8 & - \\
		& 2 (LAMA-UHN) & 20.2 & 28.2 & \textbf{31.1} & 24.7 & 24.6 & 23.0 & 27.8 & 29.5 & 21.7 \\ \midrule
		Questions w/ & 0 (original LAMA) & 32.0 & 28.5 & \textbf{35.8} & 30.4 & 32.0 & 33.1 & 28.2 & 34.9 & - \\
		KnowBert & 1 (string match filter) & 24.8 & 28.6 & \textbf{32.0} & 25.7 & 25.9 & 27.0 & 28.3 & 31.5 & - \\
		subject only & 2 (LAMA-UHN) & 22.8 & 27.7 & \textbf{30.6} & 24.9 & 25.1 & 25.5 & 27.4 & \textbf{30.6} & - \\ \bottomrule
	\end{tabularx}
	\caption{Mean Hits@1 on LAMA-Google-RE and LAMA-T-REx combined. 0: original LAMA dataset \cite{petroni2019language}, 1: after string match filter, 2: after string match filter and person name filter (LAMA-UHN). ``Questions w/ KnowBert subject only'': Evaluating on questions whose gold subject is in the KnowBert entity vocabulary. Results for K-Adapter are calculated from \citet[Table 5]{wang2020k}. See Appendix for individual relations.}
	\tablabel{lama}
\end{table*}

\subsection{LAMA-UHN}
\seclabel{filtering_lama}
Imagine a person who claims to know a lot of facts.
During a quiz, you ask them about the native language of actor Jean Marais.
They correctly answer ``French.''  
For a moment you are impressed, until you realize
that Jean is a typical French name.  So you ask the same
question about Daniel Ceccaldi (a French actor with an Italian-sounding name).
This time, the person says ``Italian.''
 
If this quiz were a QA benchmark, the person would have achieved a
respectable Hits@1 score of 50\%.
Yet, you doubt that they really \textit{knew} the first answer.

Qualitative inspection of BERT's answers to LAMA suggests that the model often behaves less like a KB and more like the person just described.
In \tabref{frenchpoets} for instance, BERT predicts native languages that are plausible for people's names, even when there is no factual basis for these predictions.
This kind of name-based reasoning is a useful strategy for getting a high score on
LAMA, as the correct answer and the best name-based guess tend to coincide
(e.g., people with Italian-sounding names frequently speak Italian).
Hence, LAMA in its current form cannot differentiate whether a model is good at reasoning about
(the surface form of) entity names, good at memorizing facts, or both.
To quantify the effect, we create LAMA-UHN (UnHelpful
Names), a subset of LAMA where overly helpful entity names are heuristically deleted:

\paragraph{Heuristic 1 (string match filter).}
We first delete all KB triples (questions) where the correct answer
(e.g., \textit{Apple}) is a case-insensitive substring of
the subject entity name (e.g., \textit{Apple Watch}).  This
affects 12\% of all triples, and up to 81\% for individual relations (see \tabref{filtered}, top).

\paragraph{Heuristic 2 (person name filter).}
Entity names can be revealing in ways that are more subtle than string matches.
As illustrated by our \textit{Jean Marais} example, a person's name can be a useful prior for guessing their native language and by extension, their nationality, place of birth, etc.
We therefore use cloze-style questions
to elicit name associations inherent in BERT, and delete
triples that correlate with them.

The heuristic is best explained via an example.
Consider again (\textbf{Jean\_Marais}, \textbf{native-language},
\textbf{French}). 
We whitespace-tokenize the subject's surface form \textit{Jean Marais} into \textit{Jean} and \textit{Marais}.
If BERT considers either name to be a common French name, then a correct answer is insufficient evidence for factual knowledge about the entity \textbf{Jean\_Marais}.
On the other hand, if neither \textit{Jean} nor \textit{Marais} are considered French, but a correct answer is given regardless, we consider it sufficient evidence of factual knowledge.

We query BERT with ``[X] is a
common name in the following language: [MASK].''
for [X] = \textit{Jean} and [X] =
\textit{Marais}.
(Depending on the relation, we replace ``language'' with ``city'' or ``country''.)
If \textit{French} is among the top-3 answers for either question, we delete the original triple.
We apply this heuristic to T-REx:P19 (place of birth), T-REx:P20 (place of death), T-REx:P27 (nationality), T-REx:P103 (native language), T-REx:P1412 (language used), Google-RE:place-of-death and Google-RE:place-of-birth.
See \tabref{filtered} (bottom) for examples and statistics.

\begin{figure}
\centering
\includegraphics[width=.99\columnwidth, trim={3.5mm 4mm 4mm 2mm}, clip]{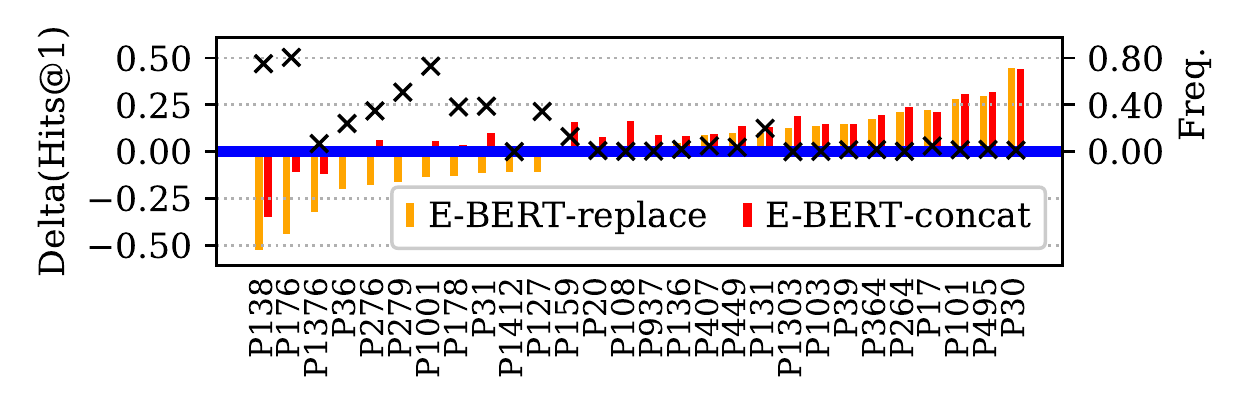}
\caption{Left y-axis (bars): delta in mean Hits@1 relative to BERT on individual LAMA relations. Right y-axis (crosses): frequency of questions where the answer is a substring of the subject entity name (i.e., questions that would be deleted by the string match filter). Model size: \texttt{BASE}. Due to space constraints, we only show relations with max absolute delta $\geq 0.075$.}
\figlabel{rel}
\end{figure}

\subsection{Results and discussion}
\tabref{lama} shows mean Hits@1 on the original LAMA dataset (0), after applying the string match filter (1), and after applying both filters (2, LAMA-UHN).
E-BERT-concat$_\mathrm{BASE}$ sets a new state of the art on LAMA, with major gains over original BERT.

To understand why, compare the performances of BERT and E-BERT-replace on LAMA-UHN:
While BERT drops by about 8\% between original LAMA and LAMA-UHN, E-BERT-replace drops by less than 1\%. 
This suggests that BERT's performance on original LAMA is partly due to the exploitation of helpful entity names, while that of E-BERT-replace is more strongly due to factual knowledge.
Since E-BERT-concat has access to entity names \textit{and} entity vectors, it can leverage and combine these complementary sources of information.
Interestingly, ERNIE and KnowBert have access to both sources too, but they do not achieve the same performance as E-BERT-concat.

\begin{figure}
\centering
\includegraphics[width=.95\columnwidth, trim={4mm 4.5mm 13mm 3.5mm}, clip]{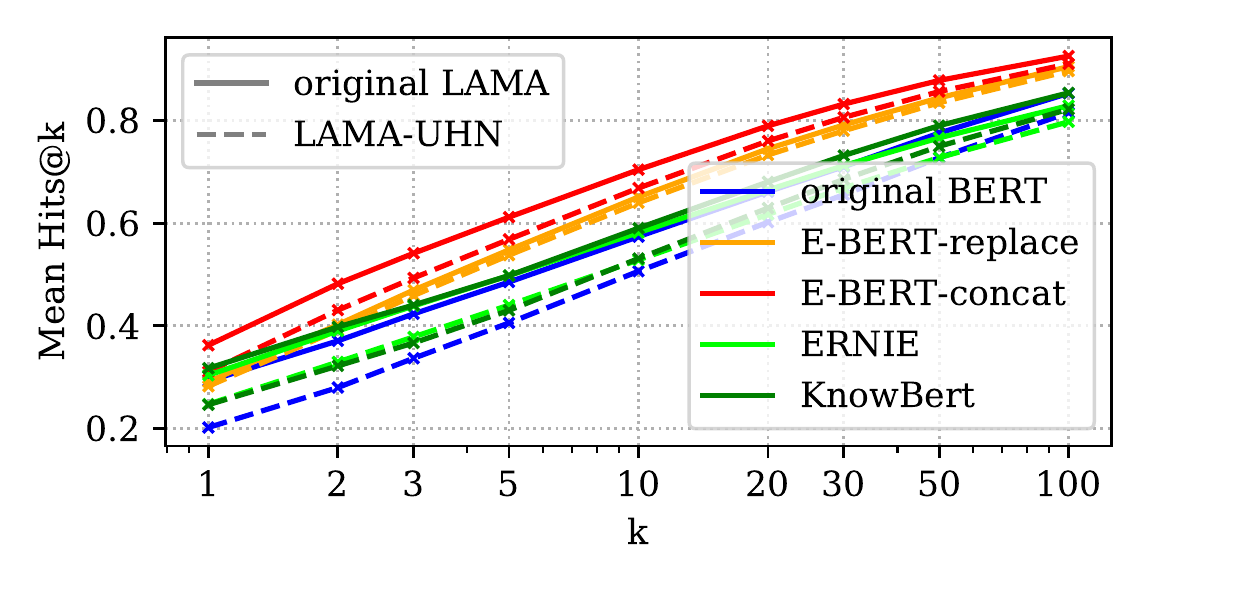}
\caption{Mean Hits@k for different $k$. Model size: \texttt{BASE}. The x-axis is on a logarithmic scale.}
\figlabel{pk}
\end{figure}

For a more in-depth analysis, \figref{rel} shows Delta(Hits@1) w.r.t. BERT (bars, left axis) on individual LAMA relations, along with the frequency of questions whose correct answer is a substring of the subject name (crosses, right axis).
The losses of E-BERT-replace are almost exclusively on relations with a high frequency of ``easy'' substring answers, while its gains are on relations where such answers are rare.
E-BERT-concat mitigates most of the losses of E-BERT-replace while keeping most of its gains.

\figref{pk} shows that the gains of E-BERT-concat over BERT, KnowBert and ERNIE in terms of mean Hits@k are especially big for $k>1$.
This means that while E-BERT-concat is moderately better than the baselines at giving the correct answer, it is a lot better at ``almost giving the correct answer''.
\citet{petroni2019language} speculate that even when factual knowledge is not salient enough for a top-1 answer, it may still be useful when finetuning on a downstream task.

\section{Downstream tasks}
\seclabel{downstream}
We now demonstrate how to use E-BERT on two downstream tasks: relation classification (RC) and entity linking (EL).
In both experiments, we keep the embedding layer ($\mathcal{E}_\mathrm{BERT}$ and/or $\mathcal{E}_\mathrm{E\text{-}BERT}$) fixed but finetune all other encoder parameters.
We use the BERT$_\mathtt{BASE}$ architecture throughout.

\subsection{Relation classification}
\seclabel{re}
In relation classification (RC), a model learns to predict the directed relation of entities $a_\mathrm{sub}$ and $a_\mathrm{obj}$ from text.
For instance, given the sentence 
\begin{itemize}[leftmargin=*]
\small
\item[]\textit{Taylor was later part of the ensemble cast in MGM 's classic World War II drama `` Battleground '' ( 1949 ) .}
\end{itemize}
with surface forms \textit{Battleground} and \textit{World War II} referring to $a_\mathrm{sub}=$ \textbf{Battleground\_(film)} and $a_\mathrm{obj}=$ \textbf{Word\_War\_II}, the model should predict the relation \textbf{primary-topic-of-work}.
We have three ways of embedding this example:
\begin{description}[leftmargin=*,font=\normalfont]
\small
\item[original BERT (wordpieces):]  \textit{[...] classic World War II drama `` Battle \#\#ground '' ( 1949 ) .}
\item[E-BERT-concat:] \textit{[...] classic} \textbf{World\_War\_II} \textit{/ World War II drama ``} \textbf{Battleground\_(film)} \textit{/ Battle \#\#ground '' ( 1949 ) .}
\item[E-BERT-replace:] \textit{[...] classic} \textbf{World\_War\_II} \textit{drama ``} \textbf{Battleground\_(film)} \textit{'' ( 1949 ) .}
\end{description}
As before, entity IDs (\textbf{bold}) are embedded by $\mathcal{E}_\mathrm{E\text{-}BERT}$ and wordpieces (\textit{italics}) by $\mathcal{E}_\mathrm{BERT}$.

\paragraph{Baselines: Wikipedia2Vec-BERT.}
To assess the impact of vector space alignment, we train two additional models (Wikipedia2Vec-BERT-concat and Wikipedia2Vec-BERT-replace) that feed non-aligned Wikipedia2Vec vectors directly into BERT (i.e., they use $\mathcal{E}_\mathrm{Wikipedia}$ instead of $\mathcal{E}_\mathrm{E\text{-}BERT}$ to embed entity IDs).

\paragraph{Data.}
We evaluate on a preprocessed dataset from \newcite{zhang2019ernie}, which is a subset of the FewRel corpus \cite{sun2018open} (see Appendix for details).
We use the FewRel oracle entity IDs, which are also used by ERNIE.
Our entity coverage is lower than ERNIE's (90\% vs. 96\%), which should put us at a disadvantage.
See Appendix for details on data and preprocessing.

\paragraph{Modeling and hyperparameters.}
We adopt the setup and hyperparameters of \newcite{zhang2019ernie}:
We use the \textit{\#} and \textit{\$} tokens to mark subject and object spans in the input, and we feed the last contextualized vector of the \textit{[CLS]} token into a randomly initialized softmax classifier.
Like \citet{zhang2019ernie}, we use the default AdamW optimizer \cite{loshchilov2018fixing} with a linear learning rate scheduler (10\% warmup) and a batch size of 32.
We tune the number of training epochs and the peak learning rate on the same parameter ranges as \citet{zhang2019ernie}. 
See Appendix for more details and an expected maximum performance plot.

\begin{table}
\scriptsize
\setlength{\tabcolsep}{.8mm}
\centering
\begin{tabularx}{.99\columnwidth}{X|ccc|ccc}
\toprule
& \multicolumn{3}{c|}{dev set} & \multicolumn{3}{c}{test set} \\ \midrule
& P & R & F1 & P & R & F1  \\
\midrule
original BERT & 85.88 & 85.81 & 85.75 & 85.57 & 85.51 & 85.45  \\
E-BERT-concat & \textbf{88.35} & \textbf{88.29} & \textbf{88.19} & \textbf{88.51} & \textbf{88.46} & \textbf{88.38} \\
E-BERT-replace  & 87.24 & 87.15 & 87.09 & 87.34 &87.33 & 87.22 \\ \midrule
Wikipedia2Vec-BERT-concat & 85.96 & 85.71 & 85.69 & 85.94 & 85.93 & 85.84  \\
Wikipedia2Vec-BERT-replace & 77.25 & 77.11 & 77.07 & 77.63 & 77.52 & 77.45 \\ \midrule
ERNIE \cite{zhang2019ernie} & - & - & - & \textbf{88.49} &\textbf{88.44} &\textbf{88.32}  \\
\bottomrule
\end{tabularx}
\caption{RC macro precision, recall and F1 (\%).}
\tablabel{fewrel}
\end{table}

\paragraph{Results and discussion.}
E-BERT-concat performs better than original BERT and slightly better than ERNIE (\tabref{fewrel}).
Recall that ERNIE required additional encoder pretraining to achieve this result.
Interestingly, E-BERT-replace (which is entity-only) beats original BERT (which is surface-form-only), i.e., aligned entity vectors seem to be more useful than entity names for this task.
The drop in F1 from E-BERT to Wikipedia2Vec-BERT shows the importance of vector space alignment.

\subsection{Entity linking}
\seclabel{el}
Entity linking (EL) is the task of detecting entity spans in a text and linking them to the underlying entity ID.
While there are recent advances in fully end-to-end EL \cite{broscheit2019investigating}, the task is typically broken down into three steps: (1) detecting spans that are potential entity spans, (2) generating sets of candidate entities for these spans, (3) selecting the correct candidate for each span.

For steps (1) and (2), we use KnowBert's candidate generator \cite{peters2019knowledge}, which is based on a precomputed span-entity co-occurrence table \cite{hoffart2011robust}.
Given an input sentence, the generator finds all spans that occur in the table, and annotates each with a set of candidates $A=\{a_1 \ldots a_N\}$ and prior probabilities $\{p(a_1) \ldots p(a_N)\}$.
Note that the candidates and priors are span- but not context-specific, and that the generator may over-generate.
For step (3), our model must therefore learn to (a) reject over-generated spans and (b) disambiguate candidates based on context.

\paragraph{Modeling.}
Recall that BERT was pretrained as a masked LM (MLM).
Given a wordpiece-tokenized input $X$ with $x_i = \textit{[MASK]}$, it has learned to predict a probability distribution over $\mathbb{L}_\mathrm{WP}$ to fill the masked position:
\begin{align}
	& \forall w \in \mathbb{L}_\mathrm{WP} \eqlabel{probwordpieces} \\ \nonumber
	& p(w | X) \propto \mathrm{exp}(\mathbf{e}_w \cdot \mathcal{F}_\mathrm{MLM}(\mathbf{h}_i) + b_w)
\end{align}
where $\mathbf{h}_i$ is the contextualized embedding of \textit{[MASK]}, $b_w$ is a learned bias and $\mathbf{e}_w = \mathcal{E}_\mathrm{BERT}(w)$.
Since $\mathcal{E}_\mathrm{E\text{-}BERT}[\mathbb{L}_\mathrm{Ent}]$ is aligned with $\mathcal{E}_\mathrm{BERT}[\mathbb{L}_\mathrm{WP}]$, the pretrained MLM should have a good initialization for predicting entities from context as well.

Based on this intuition, our \textbf{E-BERT-MLM} model repurposes the MLM for the entity selection step.
Given a wordpiece-tokenized span $s_1 \ldots s_{T_s}$ with left context $l_1 \ldots l_{T_l}$, right context $r_1 \ldots r_{T_r}$, candidates $A$ and priors $p(a)$, we define:
$$
X = l_1 \ldots l_{T_l} \text{ }\textit{[E-MASK]} \textit{ / } s_1 \ldots s_{T_s} \textit{* } r_1 \ldots r_{T_r}
$$
All tokens in $X$ except \textit{[E-MASK]} are embedded by $\mathcal{E}_\mathrm{BERT}$.
\textit{[E-MASK]} is embedded as $\frac{1}{|A|} \sum_{a \in A} \mathcal{E}_\mathrm{E\text{-}BERT}(a)$, to inform the encoder about its options for the current span. 
(See \tabref{el} for an ablation with the standard \textit{[MASK]} token.)

The output probability distribution for \textit{[E-MASK]} is not defined over $\mathbb{L}_\mathrm{WP}$ but over $A \cup \{\epsilon\}$, where $\epsilon$ stands for rejected spans (see below):
\begin{align}
	& \forall a \in A \cup \{\epsilon\} \eqlabel{probentity} \\ \nonumber
		  & p(a | X) \propto \mathrm{exp}(\mathbf{e}_a \cdot \mathcal{F}_\mathrm{MLM}(\mathbf{h}_{T_l + 1}) + b_a)
\end{align}
For all $a \in A$, $\mathbf{e}_a = \mathcal{E}_\mathrm{E\text{-}BERT}(a)$ and $b_a = \mathrm{log}(p(a))$.\footnote{To understand why we set $b_a = \mathrm{log}(p(a))$, assume that the priors are implicitly generated as $p(a) = \mathrm{exp}(b_a) / Z$, with $Z = \sum_{a'} \mathrm{exp}(b_{a'})$. It follows that $b_a = \mathrm{log}(p(a)) + \mathrm{log}(Z)$.
Since $\mathrm{log}(Z)$ is the same for all $a'$, and the softmax function is invariant to constant offsets, we can drop $\mathrm{log}(Z)$ from \eqref{probentity}.} The null-entity $\epsilon$ has parameters $\textbf{e}_\epsilon, b_\epsilon$ that are trained from scratch.

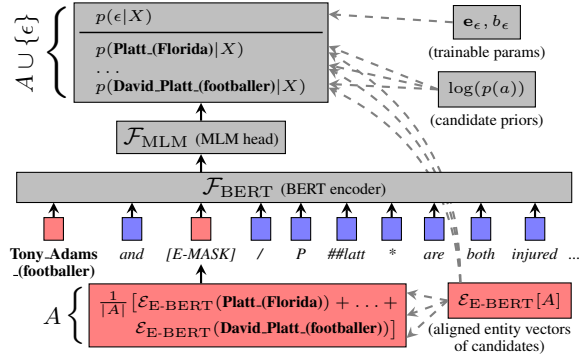
\begin{figure}[t!]
\centering
\begin{tikzpicture}[
  aligned/.style={%
    text height=.75ex,
    text depth=.25ex,
    text centered
  }
]
\newcommand{\dashrule}[1][black]{%
  \color{#1}\rule[\dimexpr.5ex-.2pt]{4pt}{.4pt}\xleaders\hbox{\rule{4pt}{0pt}\rule[\dimexpr.5ex-.2pt]{4pt}{.4pt}}\hfill\kern0pt%
}

\node (w1) [aligned] at (0,0.1) {\tiny \textbf{Tony\_Adams}};
\node [below=-1mm of w1, inner sep = 0] {\tiny \textbf{\_(footballer)}};
\node (w2) [aligned, right=.2mm of w1] {\tiny \textit{and}};
\node (w3) [aligned, right=.2mm of w2] {\tiny \textit{[E-MASK]}};
\node (w4) [aligned, right=.2mm of w3] {\tiny \textit{/}};
\node (w5) [aligned, right=1.5mm of w4] {\tiny \textit{P}};
\node (w6) [aligned, right=.2mm of w5] {\tiny \textit{\#\#latt}};
\node (w7) [aligned, right=.2mm of w6] {\tiny \textit{*}};
\node (w8) [aligned, right=1mm of w7] {\tiny \textit{are}};
\node (w9) [aligned, right=.2mm of w8] {\tiny \textit{both}};
\node (w10) [aligned, right=-.5mm of w9] {\tiny \textit{injured}};
\node (w11) [aligned, right=0mm of w10, inner sep=0pt] {\tiny \textit{...}};

\node at (0, 2.65) (bla) {};
\node at (2, 0) (xxx) {};
\node [rectangle, draw, above=1.8 of w3, fill=gray!50, inner sep=2pt]  (probs) 
{\tiny \begin{tabular}{l} $p(\epsilon|X)$ \\ \midrule $p(\textbf{Platt\_(Florida)}|X)$ \\ $\ldots$ \\ $p(\textbf{David\_Platt\_(footballer)}|X)$ \end{tabular}};

\node at (6,-.5) [rectangle, draw, fill=red!50] (ebert) {\tiny $\mathcal{E}_\mathrm{E\text{-}BERT}[A]$};
\node at (5.7, 2.3) [rectangle, draw, fill=gray!50] (bias) {\tiny $\mathrm{log}(p(a))$};
\foreach \i in {2.5mm, 5mm, 7.5mm}{
\draw [->, >=stealth, thick, dashed, gray] (bias.west) to ([yshift=\i] probs.south east); 
}
\foreach \i in {2mm, 4.5mm, 6.7mm}{
\draw [->, >=stealth, thick, dashed, gray, bend angle = 30, bend right] ([xshift=-6.5mm] ebert.north) to ([yshift=\i] probs.south east);
}


\node at (0, .45) (e1) [rectangle, minimum height=3mm, draw, fill = red!50] {};
\foreach \i in {2,4,5,...,10} {
\node at (e1-|w\i) [rectangle, minimum height = 3mm, draw, draw, fill=blue!50] (e\i) {};
}

\node at (e1-|w3) [rectangle, minimum height = 3mm, draw, draw, fill=red!50] (e3) {};
\node [draw, fill = gray!50, above = 11.2 mm of w3, inner sep=2.5pt] (mlm) {\small $\mathcal{F}_\mathrm{MLM}$ \tiny (MLM head)};

\node (lower) at (1, .8) {};
\node (upper) at (1, 1.2) {};

\node (emask) [rectangle, align=left, below=1.5mm of w4.south west, draw, fill=red!50, inner sep=3pt] 
	{\tiny $\frac{1}{|A|} \big[\mathcal{E}_\mathrm{E\text{-}BERT}(\textbf{Platt\_(Florida)}) + \ldots +$ \\[-3.5pt]
	\tiny \hfill $\mathcal{E}_\mathrm{E\text{-}BERT}(\textbf{David\_Platt\_(footballer)})\big]$};

\draw [draw, fill=gray!50] ([xshift=-10pt] e1.west |- lower) rectangle (ebert.east |- upper) node [pos=.5] {\small $\mathcal{F}_\mathrm{BERT}$ \tiny (BERT encoder)};

\node at (6.2, 1.8) (tmp) {};
\node at (6.2, .8) (tmp2) {};
\node at (6.2, 1.4) (tmp3) {};
\node at (6.2, 1.2) (tmp4) {};
\node [left=-1mm of probs] (probsbracket) {\Large $\bigg\{$};
	\node [left=-1mm of probsbracket, inner sep = 0] {\rotatebox{90}{\small $A \cup \{\epsilon\}$}};
\node [above = 4.2mm of bias, rectangle, draw, fill=gray!50] (nil) {\tiny $\mathbf{e}_\epsilon, b_\epsilon$};
\node [below=0.5mm of ebert.south east, anchor = north east, align=center, inner sep=0] {\tiny (aligned entity vectors\\[-2.5mm] \tiny of candidates)};
\foreach \i in {1,...,10} {
\draw [->, >=stealth, thick] (e\i) to (tmp2-|w\i);
}

\draw [->, >=stealth, thick] (tmp4-|w3) to (mlm.south);
\draw [->, >=stealth, thick] (mlm.north) to (probs.south-|w3);
\draw [->, >=stealth, thick] (emask.north-|w3.south) to ([yshift=3pt] w3.south);
\foreach \i in {-3mm, 0mm, 3mm}{
\draw [->, >=stealth, thick, dashed, gray] (ebert.west) to ([yshift=\i] emask.east);
}

\draw [->, >=stealth, thick, dashed, gray] (nil.west) to ([yshift=11.75mm] probs.south east);
\node [below=-.5mm of bias.south] {\tiny (candidate priors)};
\node [below=-.5mm of nil.south] {\tiny (trainable params)};
\node [left=-1mm of emask] (emaskbracket) {\large $\Big\{$};
\node [left=-1mm of emaskbracket, inner sep=0] {\small $A$};
\end{tikzpicture}
\caption{Schematic depiction of E-BERT-MLM. \textcolor{blue}{Blue}: $\mathcal{E}_\mathrm{BERT}$ wordpiece vectors. \textcolor{red}{Red}: $\mathcal{E}_\mathrm{E\text{-}BERT}$ entity vectors. The candidates $A$ and their priors $p(a)$ are from the candidate generator. Assume that the entity \textbf{Tony\_Adams\_(footballer)} was decoded in a previous iteration (see ``Iterative refinement'').}
\figlabel{schematic-el}
\end{figure}

\begin{table}[t!]
\scriptsize
\setlength{\tabcolsep}{5.5pt}
\centering
\begin{tabularx}{.99\columnwidth}{l|X|cc|cc}
\toprule
\multicolumn{2}{l|}{} & \multicolumn{2}{c|}{AIDA-A (dev)} & \multicolumn{2}{c}{AIDA-B (test)} \\
\multicolumn{2}{l|}{} & Micro & Macro & Micro & Macro \\
\midrule

\multicolumn{2}{l|}{E-BERT-MLM} & \textbf{90.8} & \textbf{89.1} & \textbf{85.0} & \textbf{84.2} \\
\multicolumn{1}{l}{} & w/o iterative refinement & 90.6 & 89.0 & - & - \\
\multicolumn{1}{l}{} & w/ standard \textit{[MASK]} token &90.3 & 88.8 & - & - \\
\multicolumn{2}{l|}{Wikipedia2Vec-BERT-MLM} & 88.7 & 86.4 & 80.6 & 81.0 \\
\multicolumn{2}{l|}{Wikipedia2Vec-BERT-random} & 88.2 & 86.1& 80.5 & 81.2 \\
\midrule
\multicolumn{2}{l|}{\citet{kolitsas2018end}} & 89.4 & 86.6 &  82.4 & 82.6  \\
\multicolumn{2}{l|}{\citet{broscheit2019investigating}} & 86.0 & - & 79.3 & - \\
\multicolumn{2}{l|}{KnowBert \cite{peters2019knowledge}} & 82.1 & - & 73.7 & - \\
\multicolumn{2}{l|}{\citet{chen2019yelm}$^\dagger$} & \textbf{92.6} & \textbf{93.6} & \textbf{87.5} & \textbf{87.7} \\
\bottomrule
\end{tabularx}
\caption{F1 (\%) on AIDA after finetuning. $^\dagger$Might not be comparable: \citet{chen2019yelm} evaluate on in-vocabulary entities only, without ensuring (or reporting) the vocabulary's coverage of the AIDA data.}
\tablabel{el}
\end{table}

\paragraph{Finetuning.}
We finetune E-BERT-MLM on the training set to minimize $\sum_{(X, \hat{a})} - \mathrm{log}(p(\hat{a}|X))$, where $(X, \hat{a})$ are pairs of potential spans and their gold entities.
If $X$ has no gold entity (if it was over-generated), then $\hat{a} = \epsilon$.\footnote{If $\hat{a} \neq \epsilon \land \hat{a} \not\in A$, we remove the span from the training set. We do \textbf{not} do this at test time, i.e., we evaluate on all gold standard entities.}

\paragraph{Iterative refinement.}
We found it useful to iteratively refine predictions during inference, similar to techniques from non-autoregressive Machine Translation \cite{ghazvininejad2019constant}.
We start with a wordpiece-tokenized input, e.g.:
\begin{itemize}[leftmargin=*]
\small
\item[] \textit{Adams and P \#\#latt are both injured and will miss England 's opening World Cup qualifier ...}
\end{itemize}
We make predictions for all potential spans that the candidate generator finds in the input.
We gather all spans with $\mathrm{argmax}_a [p(a|X)] \neq \epsilon$, sort them by $1-p(\epsilon|X)$ and replace the top-$k$\footnote{$k = \mathrm{ceil}(\frac{j(m+n)}{J}) - m$, where $1 \leq j \leq J$ is the current iteration, $m$ is the number of already decoded entities from previous iterations, and $n=|\{X: \mathrm{argmax}_a[p(a|X)] \neq \epsilon\}|$.}
non-overlapping spans with the predicted entity.
Our previous example might be partially decoded as:
\begin{itemize}[leftmargin=*]
\small
\item[] \textbf{Tony\_Adams\_(footballer)} \textit{and P \#\#latt are both injured and will miss England 's opening} \textbf{1998\_FIFA\_World\_Cup} \textit{qualifier ...}
\end{itemize}
In the next iteration, decoded entities (\textbf{bold}) are represented by $\mathcal{E}_\mathrm{E\text{-}BERT}$ in the input, while non-decoded spans continue to be represented by $\mathcal{E}_\mathrm{BERT}$ (see \figref{schematic-el}).
We set the maximum number of iterations to $J=3$, as there were no improvements beyond that point on the dev set.

\paragraph{Baselines: Wikipedia2Vec-BERT.}
We train two additional baseline models that combine BERT and Wikipedia2Vec without vector space alignment:
\begin{description}
 \setlength\itemsep{0mm}
\item[Wikipedia2Vec-BERT-MLM:] BERT and its pretrained MLM head, finetuned to predict non-aligned Wikipedia2Vec vectors. In practice, this means replacing $\mathcal{E}_\mathrm{E\text{-}BERT}$ with $\mathcal{E}_\mathrm{Wikipedia}$ in \eqref{probentity}. Embedding the \textit{[E-MASK]} token with non-aligned $\mathcal{E}_\mathrm{Wikipedia}$ led to a drop in dev set micro F1, therefore we report this baseline with the standard \textit{[MASK]} token.
\item[Wikipedia2Vec-BERT-random:] Like Wikipe-dia2Vec-BERT-MLM, but the MLM head is replaced by a randomly initialized layer.
\end{description}

\paragraph{Data.}
We train and evaluate on AIDA, a news dataset annotated with Wikipedia URLs \cite{hoffart2011robust}.
To ensure coverage of the necessary entities, we include all gold entities and all generator candidates in the entity vocabulary $\mathbb{L}_\mathrm{Ent}$, even if they fall under the Wikipedia2Vec link threshold (see \secref{systems}).
While this is based on the unrealistic assumption that we know the contents of the test set in advance, it is necessary for comparability with \citet{peters2019knowledge}, \citet{kolitsas2018end} and \citet{broscheit2019investigating}, who also design their entity vocabulary around the data.
See Appendix for more details on data and preprocessing.
We evaluate strong match F1, i.e., a prediction must have the same start, end and entity (URL) as the gold standard.
URLs that redirect to the same Wikipedia page are considered equivalent.

\paragraph{Hyperparameters.}
We train for 10 epochs with the AdamW optimizer \cite{loshchilov2018fixing} and a linear learning rate scheduler (10\% warmup), and we select the best epoch on the dev set.
We tune peak learning rate and batch size on the dev set (see Appendix).

\paragraph{Results and discussion.}
\tabref{el} shows that
E-BERT-MLM is competitive with previous work on AIDA. 
The aligned entity vectors play a key role in this performance, as they give the model a good initialization for predicting entities from context.
When we remove this initialization by using non-aligned entity vectors (Wikipedia2Vec-BERT baselines), we get worse unsupervised performance (\tabref{zeroshot}), slower convergence during finetuning (\figref{convergence}), and a lower final F1 (\tabref{el}).

\section{Conclusion}
We introduced \textbf{E-BERT}, an efficient yet effective way of injecting factual knowledge about entities into the BERT pretrained Language Model.
We showed how to align Wikipedia2Vec entity vectors with BERT's wordpiece vector space, and how to feed the aligned vectors into BERT as if they were wordpiece vectors.
In doing so, we made no changes to the BERT encoder itself.
This stands in contrast to other entity-enhanced versions of BERT,
such as ERNIE or KnowBert, which add encoder layers and require expensive further pretraining.

We set a new state of the art on LAMA, a recent unsupervised QA benchmark.
Furthermore, we presented evidence that the original BERT model sometimes relies on the surface forms of entity names (rather than ``true'' factual knowledge) for this task.
To quantify this effect, we introduced LAMA-UHN, a subset of LAMA where questions with helpful entity names are deleted.

We also showed how to apply E-BERT to two supervised tasks: relation classification and entity linking.
On both tasks, we achieve competitive results relative to BERT and other baselines.

\begin{figure}[t!]
\centering
\includegraphics[width=.99\columnwidth, trim={3.5mm 0mm 2.4mm 3.3mm}, clip]{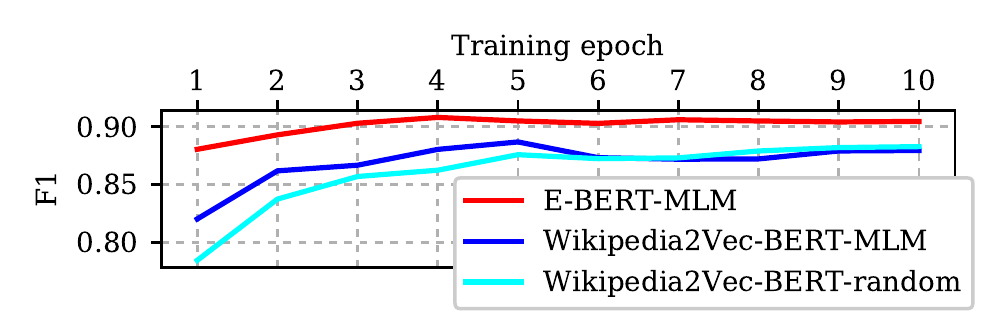}
\caption{AIDA dev set micro F1 after every epoch.}
\figlabel{convergence}
\end{figure}

\begin{table}[t!]
\centering
\scriptsize
\setlength{\tabcolsep}{1.8mm}
\newcommand{\mysep}{\hfill / \hfill}
\begin{tabularx}{.99\columnwidth}{lX|rrr}
	\toprule
	& & \multicolumn{1}{c}{P} & \multicolumn{1}{c}{R} & \multicolumn{1}{c}{F1} \\ \midrule 
	\multicolumn{2}{l|}{E-BERT-MLM} & 21.1 & 61.8 & 31.5 \\
		   & w/ standard \textit{[MASK]} token & 23.3 & 65.2 & 34.3 \\ 
	\multicolumn{2}{l|}{Wikipedia2Vec-BERT-MLM} & 1.3 & 8.3 & 2.3 \\
\multicolumn{2}{l|}{Wikipedia2Vec-BERT-random} & 1.3 & 6.8 & 2.2 \\
	\bottomrule
\end{tabularx}
\caption{Unsupervised AIDA dev set micro precision / recall / F1 (\%), before finetuning. Results are without iterative refinement.}

\tablabel{zeroshot}
\end{table}

\bibliography{main}
\bibliographystyle{acl_natbib}

\clearpage

\appendix
{\large \noindent \textbf{E-BERT: Efficient-Yet-Effective Entity Embeddings for BERT (Appendix)}}

\section*{Unsupervised QA (LAMA)}
\subsection*{Data}
We downloaded the LAMA dataset from \url{https://dl.fbaipublicfiles.com/LAMA/data.zip}.
We use the LAMA-T-\mbox{REx} and LAMA-Google-RE relations, which are aimed at factual knowledge.
\tabref{lama-stats} shows results on indiviual relations, as well as the number of questions per relation before and after applying the LAMA-UHN heuristics.

\subsection*{Preprocessing}
As mentioned in \secref{lama}, we do not use LAMA's oracle entity IDs.
Instead, we map surface forms to entity IDs via the Wikidata query API (\url{https://query.wikidata.org}).
For example, to look up \textit{Jean Marais}:

\begin{lstlisting}[label=lst:sparql, upquote=true, basicstyle=\ttfamily\small,frame=none, showstringspaces=false]
SELECT ?id ?str WHERE {
  ?id rdfs:label ?str .
  VALUES ?str { 'Jean Marais'@en } .
  FILTER((LANG(?str)) = 'en') . 
}
\end{lstlisting}

If more than one Wikidata ID is returned, we select the lowest one.
We then map Wikidata IDs to the corresponding Wikipedia URLs:

\begin{lstlisting}[label=lst:sparql, upquote=true, basicstyle=\ttfamily\small,frame=none, showstringspaces=false]
SELECT ?id ?wikiurl WHERE {
  VALUES ?id { wd:Q168359 } .
  ?wikiurl schema:about ?id .
  ?wikiurl schema:inLanguage 'en' .
  FILTER REGEX(str(?wikiurl), 
  	'.*en.wikipedia.org.*') .
}
\end{lstlisting}

\section*{Relation classification}
\subsection*{Data}
The RC dataset, which is a subset of the FewRel corpus, was compiled by \citet{zhang2019ernie}. 
We downloaded it from \url{https://cloud.tsinghua.edu.cn/f/32668247e4fd4f9789f2/}.
\tabref{rc-stats} shows dataset statistics.

\subsection*{Preprocessing}
The dataset contains sentences with annotated subject and object entity mentions, their oracle entity IDs and their relation (which must be predicted).
We use the BERT wordpiece tokenizer to tokenize the sentence and insert special wordpieces to mark the entity mentions: \textit{\#} for subjects and \textit{\$} for objects.
Then, we insert the entity IDs.
For example, an input to E-BERT-concat would look like this:
\begin{itemize}
\small
\item[] \textit{[CLS] Taylor was later part of the ensemble cast in MGM 's classic \$ \textbf{World\_War\_II} / World War II \$ drama `` \# \textbf{Battleground\_(film)} / Battle \#\#ground \# '' ( 1949 ) . [SEP]}
\end{itemize}
We use the oracle entity IDs of the dataset, which are also used by ERNIE \cite{zhang2019ernie}.

\subsection*{Hyperparameters}
We tune peak learning rate and number of epochs on the dev set (selection criterion: macro F1).
We do a full search over the same hyperparameter space as \citet{zhang2019ernie}:
\begin{description}
\setlength\itemsep{0mm}
\item[Learning rate:] $[2\cdot 10^{-5}, 3\cdot 10^{-5}, \mathbf{5 \cdot 10^{-5}}]$
\item[Number of epochs:] $[3,4,5,6,7,8,9,\mathbf{10}]$
\end{description}
The best configuration for E-BERT-concat is marked in bold.
\figref{dodge_fewrel} shows expected maximum performance as a function of the number of evaluated configurations \cite{dodge2019show}.

\section*{Entity linking (AIDA)}
\subsection*{Data}
We downloaded the AIDA dataset from:
\begin{itemize}
\setlength\itemsep{0mm}
\item \url{https://allennlp.s3-us-west-2.amazonaws.com/knowbert/wiki_entity_linking/aida_train.txt}
\item \url{https://allennlp.s3-us-west-2.amazonaws.com/knowbert/wiki_entity_linking/aida_dev.txt}
\item \url{https://allennlp.s3-us-west-2.amazonaws.com/knowbert/wiki_entity_linking/aida_test.txt}
\end{itemize}

\subsection*{Preprocessing}
Each AIDA file contains documents with annotated entity spans (which must be predicted).
The documents are already whitespace tokenized, and we further tokenize words into wordpieces with the standard BERT tokenizer.
If a document is too long (length $>512$), we split it into smaller chunks by (a) finding the sentence boundary that is closest to the document midpoint, (b) splitting the document, and (c) repeating this process recursively until all chunks are short enough.
\tabref{el-stats} shows dataset statistics.

\subsubsection*{Hyperparameters}
We tune batch size and peak learning rate on the AIDA dev set (selection criterion: strong match micro F1).
We do a full search over the following hyperparameter space:
\begin{description}
\setlength\itemsep{0mm}
\item[Batch size:] $[16, 32, 64, \mathbf{128}]$
\item[Learning rate:] $[\mathbf{2\cdot 10^{-5}}, 3\cdot 10^{-5}, 5 \cdot 10^{-5}]$
\end{description}

The best configuration for E-BERT-MLM is marked in bold.
\figref{dodge_aida} shows expected maximum performance as a function of the number of evaluated configurations \cite{dodge2019show}.

\newpage 

\begin{table}[h!]
\scriptsize
\centering
\begin{tabularx}{.99\columnwidth}{X|ccc}
\toprule
\# relations & 80 \\
\# unique entities & 54648 \\
\midrule
& train & dev & test \\ \cmidrule{2-4}
\# samples & 8000 & 16000 & 16000 \\
\# samples per relation & 100 & 200 & 200 \\
\bottomrule
\end{tabularx}
\centering
\caption{Relation classification dataset statistics.}
\tablabel{rc-stats}
\end{table}

\begin{table}[h!]
\scriptsize
\begin{tabularx}{.99\columnwidth}{X|ccc}
\toprule
\# unique gold entities & 5574 \\
\# unique candidate entities & 463663 \\ 
\midrule
& train & dev & test \\ \cmidrule{2-4}
\# documents & 946 & 216 & 231  \\
\# documents (after chunking) & 1111 & 276 & 271 \\
\# potential spans (candidate generator) & 153103 & 38012 & 34936 \\
\# gold entities & 18454 & 4778 & 4478 \\
\bottomrule
\end{tabularx}
\centering
\caption{Entity linking (AIDA) dataset statistics.}
\tablabel{el-stats}
\end{table}

\begin{figure}[h!]
\centering
\includegraphics[width=.99\columnwidth]{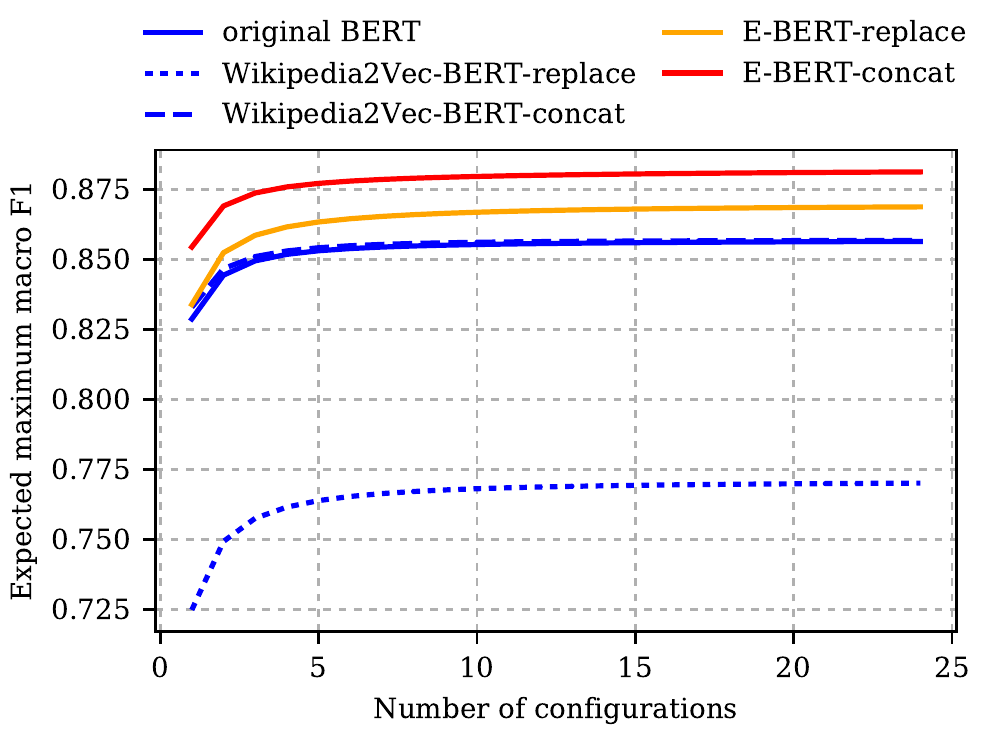}
\caption{Relation classification: Expected maximum macro F1 (dev set) as a function of the number of hyperparameter configurations.}
\figlabel{dodge_fewrel}
\end{figure}

\begin{figure}[h!]
\centering
\includegraphics[width=.99\columnwidth]{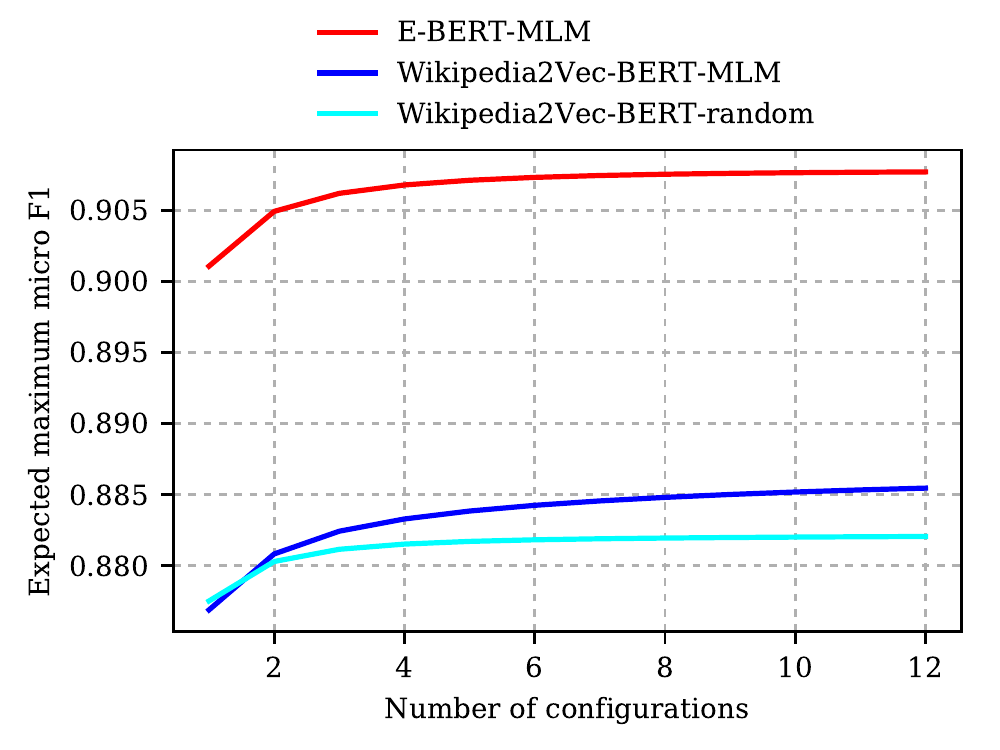}
\caption{Entity linking: Expected maximum micro F1 (dev set) as a function of the number of hyperparameter configurations.}
\figlabel{dodge_aida}
\end{figure}

\begin{table*}
\scriptsize
\centering
\begin{tabularx}{.99\textwidth}{X|rrrrr|rrr|r}
\toprule
\multicolumn{1}{r|}{Model size:} & \multicolumn{5}{c|}{\texttt{BASE}} & \multicolumn{3}{c|}{\texttt{LARGE}} \\ \midrule
\multirow{2}{*}{\diagbox{Relation (dataset)}{Model}}& original & E-BERT & E-BERT- & ERNIE & Know- & original & E-BERT- & E-BERT- & number of \\
& \multicolumn{1}{c}{BERT} & \multicolumn{1}{c}{replace} & \multicolumn{1}{c}{concat} & & \multicolumn{1}{c|}{Bert} & \multicolumn{1}{c}{BERT} & \multicolumn{1}{c}{replace} & \multicolumn{1}{c|}{concat} & \multicolumn{1}{c}{questions} \\
\midrule
T-REx:P17 \hfill (0, original LAMA)            &           31.3 &            53.7 &           52.4 &   55.3 &      23.7 &                 36.5 &                  43.3 &                 42.8 &   930 \\
T-REx:P17 \hfill (1)            &           31.0 &            55.0 &           53.3 &   55.5 &      23.2 &                 36.2 &                  44.5 &                 43.3 &   885 \\
T-REx:P17 \hfill (2, LAMA-UHN)            &           31.0 &            55.0 &           53.3 &   55.5 &      23.2 &                 36.2 &                  44.5 &                 43.3 &   885 \\ \midrule
T-REx:P19 \hfill (0, original LAMA)            &           21.1 &            26.4 &           28.1 &   28.7 &      23.3 &                 22.2 &                  24.6 &                 25.3 &   944 \\
T-REx:P19 \hfill (1)            &           20.6 &            26.5 &           27.5 &   28.2 &      22.9 &                 21.8 &                  24.5 &                 24.8 &   933 \\
T-REx:P19 \hfill (2, LAMA-UHN)            &            9.8 &            20.3 &           18.7 &   19.4 &      12.2 &                 11.7 &                  18.1 &                 15.5 &   728 \\ \midrule
T-REx:P20 \hfill (0, original LAMA)            &           27.9 &            29.7 &           35.8 &   16.6 &      31.1 &                 31.7 &                  37.1 &                 33.5 &   953 \\
T-REx:P20 \hfill (1)            &           28.2 &            29.9 &           36.0 &   16.5 &      31.0 &                 32.0 &                  37.2 &                 33.8 &   944 \\
T-REx:P20 \hfill (2, LAMA-UHN)            &           15.5 &            21.5 &           23.3 &    8.4 &      20.0 &                 18.9 &                  27.3 &                 22.6 &   656 \\ \midrule
T-REx:P27 \hfill (0, original LAMA)            &            0.0 &             0.0 &            0.1 &    0.0 &       0.1 &                  0.0 &                   0.0 &                  0.1 &   966 \\
T-REx:P27 \hfill (1)            &            0.0 &             0.0 &            0.1 &    0.0 &       0.1 &                  0.0 &                   0.0 &                  0.1 &   945 \\
T-REx:P27 \hfill (2, LAMA-UHN)            &            0.0 &             0.0 &            0.2 &    0.0 &       0.1 &                  0.0 &                   0.0 &                  0.2 &   423 \\ \midrule
T-REx:P30 \hfill (0, original LAMA)            &           25.4 &            69.9 &           69.8 &   66.8 &      24.0 &                 28.0 &                  75.0 &                 60.4 &   975 \\
T-REx:P30 \hfill (1)            &           25.1 &            70.3 &           69.9 &   66.6 &      23.9 &                 27.5 &                  75.0 &                 60.3 &   963 \\
T-REx:P30 \hfill (2, LAMA-UHN)            &           25.1 &            70.3 &           69.9 &   66.6 &      23.9 &                 27.5 &                  75.0 &                 60.3 &   963 \\ \midrule
T-REx:P31 \hfill (0, original LAMA)            &           36.7 &            25.5 &           46.9 &   43.7 &      18.7 &                 30.2 &                  12.3 &                 16.1 &   922 \\
T-REx:P31 \hfill (1)            &           21.1 &            28.4 &           35.8 &   30.3 &      12.4 &                 16.3 &                   9.9 &                  9.8 &   564 \\
T-REx:P31 \hfill (2, LAMA-UHN)            &           21.1 &            28.4 &           35.8 &   30.3 &      12.4 &                 16.3 &                   9.9 &                  9.8 &   564 \\ \midrule
T-REx:P36 \hfill (0, original LAMA)            &           62.2 &            42.1 &           61.6 &   57.3 &      62.2 &                 67.0 &                  44.7 &                 66.0 &   703 \\
T-REx:P36 \hfill (1)            &           51.5 &            41.9 &           53.9 &   45.9 &      51.7 &                 57.5 &                  43.8 &                 58.8 &   534 \\
T-REx:P36 \hfill (2, LAMA-UHN)            &           51.5 &            41.9 &           53.9 &   45.9 &      51.7 &                 57.5 &                  43.8 &                 58.8 &   534 \\ \midrule
T-REx:P37 \hfill (0, original LAMA)            &           54.6 &            51.2 &           56.5 &   60.2 &      53.1 &                 61.5 &                  54.3 &                 62.7 &   966 \\
T-REx:P37 \hfill (1)            &           52.9 &            51.6 &           55.5 &   59.4 &      51.9 &                 60.5 &                  54.2 &                 62.1 &   924 \\
T-REx:P37 \hfill (2, LAMA-UHN)            &           52.9 &            51.6 &           55.5 &   59.4 &      51.9 &                 60.5 &                  54.2 &                 62.1 &   924 \\ \midrule
T-REx:P39 \hfill (0, original LAMA)            &            8.0 &            22.9 &           22.5 &   17.0 &      17.2 &                  4.7 &                   8.1 &                  8.6 &   892 \\
T-REx:P39 \hfill (1)            &            7.5 &            23.0 &           22.3 &   17.1 &      16.5 &                  4.6 &                   8.1 &                  8.5 &   878 \\
T-REx:P39 \hfill (2, LAMA-UHN)            &            7.5 &            23.0 &           22.3 &   17.1 &      16.5 &                  4.6 &                   8.1 &                  8.5 &   878 \\ \midrule
T-REx:P47 \hfill (0, original LAMA)            &           13.7 &             8.9 &           10.8 &    9.8 &      14.0 &                 18.2 &                  15.1 &                 15.9 &   922 \\
T-REx:P47 \hfill (1)            &           13.6 &             9.1 &           10.7 &    9.6 &      13.9 &                 18.6 &                  15.2 &                 15.9 &   904 \\
T-REx:P47 \hfill (2, LAMA-UHN)            &           13.6 &             9.1 &           10.7 &    9.6 &      13.9 &                 18.6 &                  15.2 &                 15.9 &   904 \\ \midrule
T-REx:P101 \hfill (0, original LAMA)           &            9.9 &            37.8 &           40.8 &   16.7 &      12.2 &                 11.5 &                  37.8 &                 36.1 &   696 \\
T-REx:P101 \hfill (1)           &            9.5 &            38.2 &           40.9 &   16.1 &      11.4 &                 10.8 &                  38.0 &                 35.8 &   685 \\
T-REx:P101 \hfill (2, LAMA-UHN)           &            9.5 &            38.2 &           40.9 &   16.1 &      11.4 &                 10.8 &                  38.0 &                 35.8 &   685 \\ \midrule
T-REx:P103 \hfill (0, original LAMA)           &           72.2 &            85.8 &           86.8 &   85.5 &      73.4 &                 78.2 &                  84.4 &                 84.9 &   977 \\
T-REx:P103 \hfill (1)           &           72.1 &            85.7 &           86.8 &   85.4 &      73.3 &                 78.2 &                  84.4 &                 84.9 &   975 \\
T-REx:P103 \hfill (2, LAMA-UHN)           &           45.8 &            81.9 &           74.7 &   83.6 &      72.2 &                 58.6 &                  81.2 &                 71.1 &   415 \\ \midrule
T-REx:P106 \hfill (0, original LAMA)           &            0.6 &             6.5 &            5.4 &    8.4 &       1.6 &                  0.6 &                   4.3 &                  2.1 &   958 \\
T-REx:P106 \hfill (1)           &            0.6 &             6.5 &            5.4 &    8.4 &       1.6 &                  0.6 &                   4.3 &                  2.1 &   958 \\
T-REx:P106 \hfill (2, LAMA-UHN)           &            0.6 &             6.5 &            5.4 &    8.4 &       1.6 &                  0.6 &                   4.3 &                  2.1 &   958 \\ \midrule
T-REx:P108 \hfill (0, original LAMA)           &            6.8 &             9.9 &           23.2 &   14.1 &      10.7 &                  1.6 &                  11.7 &                 15.9 &   383 \\
T-REx:P108 \hfill (1)           &            6.5 &             9.9 &           23.0 &   13.9 &      10.5 &                  1.3 &                  11.8 &                 16.0 &   382 \\
T-REx:P108 \hfill (2, LAMA-UHN)           &            6.5 &             9.9 &           23.0 &   13.9 &      10.5 &                  1.3 &                  11.8 &                 16.0 &   382 \\ \midrule
T-REx:P127 \hfill (0, original LAMA)           &           34.8 &            24.0 &           34.9 &   36.2 &      31.4 &                 34.8 &                  25.3 &                 35.8 &   687 \\
T-REx:P127 \hfill (1)           &           14.2 &            19.7 &           23.5 &   17.1 &      15.5 &                 14.6 &                  21.1 &                 24.6 &   451 \\
T-REx:P127 \hfill (2, LAMA-UHN)           &           14.2 &            19.7 &           23.5 &   17.1 &      15.5 &                 14.6 &                  21.1 &                 24.6 &   451 \\ \bottomrule
\end{tabularx}
\caption{Mean Hits@1 and number of questions per LAMA relation. 0: original LAMA dataset, 1: after applying heuristic 1 (string match filter), 2: after applying both heuristics (LAMA-UHN).}
\tablabel{lama-stats}
\end{table*}
\begin{table*}
\scriptsize
\centering
\begin{tabularx}{.99\textwidth}{X|rrrrr|rrr|r}
\toprule
\multicolumn{1}{r|}{Model size:} & \multicolumn{5}{c|}{\texttt{BASE}} & \multicolumn{3}{c|}{\texttt{LARGE}} \\ \midrule
\multirow{2}{*}{\diagbox{Relation (dataset)}{Model}}& original & E-BERT & E-BERT- & ERNIE & Know- & original & E-BERT- & E-BERT- & number of \\
& \multicolumn{1}{c}{BERT} & \multicolumn{1}{c}{replace} & \multicolumn{1}{c}{concat} & & \multicolumn{1}{c|}{Bert} & \multicolumn{1}{c}{BERT} & \multicolumn{1}{c}{replace} & \multicolumn{1}{c|}{concat} & \multicolumn{1}{c}{questions} \\
\midrule
T-REx:P131 \hfill (0, original LAMA)           &           23.3 &            33.4 &           36.4 &   37.3 &      27.7 &                 26.3 &                  31.4 &                 37.2 &   881 \\
T-REx:P131 \hfill (1)           &           16.7 &            32.0 &           33.9 &   32.7 &      21.5 &                 20.1 &                  31.0 &                 33.4 &   706 \\
T-REx:P131 \hfill (2, LAMA-UHN)           &           16.7 &            32.0 &           33.9 &   32.7 &      21.5 &                 20.1 &                  31.0 &                 33.4 &   706 \\ \midrule
T-REx:P136 \hfill (0, original LAMA)           &            0.8 &             5.2 &            9.1 &    0.6 &       0.6 &                  1.3 &                   6.9 &                 13.1 &   931 \\
T-REx:P136 \hfill (1)           &            0.2 &             5.1 &            8.7 &    0.2 &       0.1 &                  0.2 &                   6.9 &                 12.2 &   913 \\
T-REx:P136 \hfill (2, LAMA-UHN)           &            0.2 &             5.1 &            8.7 &    0.2 &       0.1 &                  0.2 &                   6.9 &                 12.2 &   913 \\ \midrule
T-REx:P138 \hfill (0, original LAMA)           &           61.6 &             8.8 &           26.5 &    0.2 &      63.7 &                 45.1 &                   2.6 &                 24.0 &   645 \\
T-REx:P138 \hfill (1)           &            5.0 &            10.0 &            8.8 &    0.0 &       6.9 &                  4.4 &                   4.4 &                  6.2 &   160 \\
T-REx:P138 \hfill (2, LAMA-UHN)           &            5.0 &            10.0 &            8.8 &    0.0 &       6.9 &                  4.4 &                   4.4 &                  6.2 &   160 \\ \midrule
T-REx:P140 \hfill (0, original LAMA)           &            0.6 &             0.6 &            1.1 &    0.0 &       0.8 &                  0.6 &                   1.1 &                  0.6 &   473 \\
T-REx:P140 \hfill (1)           &            0.4 &             0.6 &            0.9 &    0.0 &       0.6 &                  0.4 &                   0.9 &                  0.4 &   467 \\
T-REx:P140 \hfill (2, LAMA-UHN)           &            0.4 &             0.6 &            0.9 &    0.0 &       0.6 &                  0.4 &                   0.9 &                  0.4 &   467 \\ \midrule
T-REx:P159 \hfill (0, original LAMA)           &           32.4 &            30.3 &           48.3 &   41.8 &      36.8 &                 34.7 &                  22.3 &                 45.2 &   967 \\
T-REx:P159 \hfill (1)           &           23.1 &            31.6 &           41.9 &   34.4 &      28.7 &                 25.6 &                  20.9 &                 37.8 &   843 \\
T-REx:P159 \hfill (2, LAMA-UHN)           &           23.1 &            31.6 &           41.9 &   34.4 &      28.7 &                 25.6 &                  20.9 &                 37.8 &   843 \\ \midrule
T-REx:P176 \hfill (0, original LAMA)           &           85.6 &            41.6 &           74.6 &   81.8 &      90.0 &                 87.5 &                  36.6 &                 81.3 &   982 \\
T-REx:P176 \hfill (1)           &           31.4 &            42.9 &           51.8 &   26.2 &      51.3 &                 40.8 &                  44.5 &                 57.1 &   191 \\
T-REx:P176 \hfill (2, LAMA-UHN)           &           31.4 &            42.9 &           51.8 &   26.2 &      51.3 &                 40.8 &                  44.5 &                 57.1 &   191 \\ \midrule
T-REx:P178 \hfill (0, original LAMA)           &           62.8 &            49.8 &           66.6 &   60.1 &      70.3 &                 70.8 &                  51.2 &                 69.4 &   592 \\
T-REx:P178 \hfill (1)           &           40.7 &            42.6 &           51.6 &   36.9 &      52.2 &                 53.6 &                  51.1 &                 57.7 &   366 \\
T-REx:P178 \hfill (2, LAMA-UHN)           &           40.7 &            42.6 &           51.6 &   36.9 &      52.2 &                 53.6 &                  51.1 &                 57.7 &   366 \\ \midrule
T-REx:P190 \hfill (0, original LAMA)           &            2.4 &             2.9 &            2.5 &    2.6 &       2.8 &                  2.3 &                   2.3 &                  2.8 &   995 \\
T-REx:P190 \hfill (1)           &            1.5 &             2.4 &            1.6 &    1.6 &       2.0 &                  1.7 &                   1.9 &                  2.3 &   981 \\
T-REx:P190 \hfill (2, LAMA-UHN)           &            1.5 &             2.4 &            1.6 &    1.6 &       2.0 &                  1.7 &                   1.9 &                  2.3 &   981 \\ \midrule
T-REx:P264 \hfill (0, original LAMA)           &            9.6 &            30.5 &           33.6 &   13.3 &      21.2 &                  8.2 &                  23.1 &                 15.6 &   429 \\
T-REx:P264 \hfill (1)           &            9.6 &            30.6 &           33.4 &   13.3 &      21.3 &                  8.2 &                  23.1 &                 15.7 &   428 \\
T-REx:P264 \hfill (2, LAMA-UHN)           &            9.6 &            30.6 &           33.4 &   13.3 &      21.3 &                  8.2 &                  23.1 &                 15.7 &   428 \\ \midrule
T-REx:P276 \hfill (0, original LAMA)           &           41.5 &            23.8 &           47.7 &   48.4 &      43.3 &                 43.8 &                  23.1 &                 51.8 &   959 \\
T-REx:P276 \hfill (1)           &           19.8 &            26.1 &           31.7 &   27.0 &      20.6 &                 23.4 &                  25.0 &                 36.0 &   625 \\
T-REx:P276 \hfill (2, LAMA-UHN)           &           19.8 &            26.1 &           31.7 &   27.0 &      20.6 &                 23.4 &                  25.0 &                 36.0 &   625 \\ \midrule
T-REx:P279 \hfill (0, original LAMA)           &           30.7 &            14.7 &           30.7 &   29.4 &      31.6 &                 33.5 &                  15.5 &                 29.8 &   963 \\
T-REx:P279 \hfill (1)           &            3.8 &             8.6 &            8.0 &    4.6 &       5.3 &                  6.8 &                   8.6 &                 10.1 &   474 \\
T-REx:P279 \hfill (2, LAMA-UHN)           &            3.8 &             8.6 &            8.0 &    4.6 &       5.3 &                  6.8 &                   8.6 &                 10.1 &   474 \\ \midrule
T-REx:P361 \hfill (0, original LAMA)           &           23.6 &            19.6 &           23.0 &   25.8 &      26.6 &                 27.4 &                  22.3 &                 25.4 &   932 \\
T-REx:P361 \hfill (1)           &           12.6 &            17.9 &           17.7 &   13.7 &      15.3 &                 18.5 &                  20.2 &                 22.0 &   633 \\
T-REx:P361 \hfill (2, LAMA-UHN)           &           12.6 &            17.9 &           17.7 &   13.7 &      15.3 &                 18.5 &                  20.2 &                 22.0 &   633 \\ \midrule
T-REx:P364 \hfill (0, original LAMA)           &           44.5 &            61.7 &           64.0 &   48.0 &      40.9 &                 51.1 &                  60.6 &                 61.3 &   856 \\
T-REx:P364 \hfill (1)           &           43.5 &            61.7 &           63.5 &   47.4 &      40.0 &                 50.7 &                  60.5 &                 61.2 &   841 \\
T-REx:P364 \hfill (2, LAMA-UHN)           &           43.5 &            61.7 &           63.5 &   47.4 &      40.0 &                 50.7 &                  60.5 &                 61.2 &   841 \\ \midrule
T-REx:P407 \hfill (0, original LAMA)           &           59.2 &            68.0 &           68.8 &   53.8 &      60.1 &                 62.1 &                  57.9 &                 56.3 &   877 \\
T-REx:P407 \hfill (1)           &           57.6 &            69.5 &           67.9 &   53.1 &      58.6 &                 61.0 &                  59.0 &                 55.2 &   834 \\
T-REx:P407 \hfill (2, LAMA-UHN)           &           57.6 &            69.5 &           67.9 &   53.1 &      58.6 &                 61.0 &                  59.0 &                 55.2 &   834 \\ \midrule
T-REx:P413 \hfill (0, original LAMA)           &            0.5 &             0.1 &            0.0 &    0.0 &      41.7 &                  4.1 &                  14.0 &                  7.0 &   952 \\
T-REx:P413 \hfill (1)           &            0.5 &             0.1 &            0.0 &    0.0 &      41.7 &                  4.1 &                  14.0 &                  7.0 &   952 \\
T-REx:P413 \hfill (2, LAMA-UHN)           &            0.5 &             0.1 &            0.0 &    0.0 &      41.7 &                  4.1 &                  14.0 &                  7.0 &   952 \\ \bottomrule
\end{tabularx}
\caption{Mean Hits@1 and number of questions per LAMA relation (cont'd). 0: original LAMA dataset, 1: after applying heuristic 1 (string match filter), 2: after applying both heuristics (LAMA-UHN).}
\end{table*}
\begin{table*}
\scriptsize
\centering
\begin{tabularx}{.99\textwidth}{X|rrrrr|rrr|r}
\toprule
\multicolumn{1}{r|}{Model size:} & \multicolumn{5}{c|}{\texttt{BASE}} & \multicolumn{3}{c|}{\texttt{LARGE}} \\ \midrule
\multirow{2}{*}{\diagbox{Relation (dataset}{Model}}& original & E-BERT & E-BERT- & ERNIE & Know- & original & E-BERT- & E-BERT- & number of \\
& \multicolumn{1}{c}{BERT} & \multicolumn{1}{c}{replace} & \multicolumn{1}{c}{concat} & & \multicolumn{1}{c|}{Bert} & \multicolumn{1}{c}{BERT} & \multicolumn{1}{c}{replace} & \multicolumn{1}{c|}{concat} & \multicolumn{1}{c}{questions} \\
\midrule
T-REx:P449 \hfill (0, original LAMA)           &           20.9 &            30.9 &           34.7 &   33.8 &      57.0 &                 24.0 &                  32.5 &                 28.6 &   881 \\
T-REx:P449 \hfill (1)           &           18.8 &            31.1 &           33.4 &   32.0 &      56.0 &                 21.8 &                  32.9 &                 27.5 &   848 \\
T-REx:P449 \hfill (2, LAMA-UHN)           &           18.8 &            31.1 &           33.4 &   32.0 &      56.0 &                 21.8 &                  32.9 &                 27.5 &   848 \\ \midrule
T-REx:P463 \hfill (0, original LAMA)           &           67.1 &            61.8 &           68.9 &   43.1 &      35.6 &                 61.3 &                  52.0 &                 66.7 &   225 \\
T-REx:P463 \hfill (1)           &           67.1 &            61.8 &           68.9 &   43.1 &      35.6 &                 61.3 &                  52.0 &                 66.7 &   225 \\
T-REx:P463 \hfill (2, LAMA-UHN)           &           67.1 &            61.8 &           68.9 &   43.1 &      35.6 &                 61.3 &                  52.0 &                 66.7 &   225 \\ \midrule
T-REx:P495 \hfill (0, original LAMA)           &           16.5 &            46.3 &           48.3 &    1.0 &      30.8 &                 29.7 &                  56.7 &                 46.9 &   909 \\
T-REx:P495 \hfill (1)           &           15.0 &            46.0 &           47.5 &    0.9 &      29.6 &                 28.5 &                  56.6 &                 46.2 &   892 \\
T-REx:P495 \hfill (2, LAMA-UHN)           &           15.0 &            46.0 &           47.5 &    0.9 &      29.6 &                 28.5 &                  56.6 &                 46.2 &   892 \\ \midrule
T-REx:P527 \hfill (0, original LAMA)           &           11.1 &             7.4 &           11.9 &    5.4 &      12.9 &                 10.5 &                   8.9 &                 12.9 &   976 \\
T-REx:P527 \hfill (1)           &            5.7 &             7.6 &            8.7 &    0.5 &       3.0 &                  4.2 &                   8.7 &                  6.3 &   804 \\
T-REx:P527 \hfill (2, LAMA-UHN)           &            5.7 &             7.6 &            8.7 &    0.5 &       3.0 &                  4.2 &                   8.7 &                  6.3 &   804 \\ \midrule
T-REx:P530 \hfill (0, original LAMA)           &            2.8 &             1.8 &            2.0 &    2.3 &       2.8 &                  2.7 &                   2.3 &                  2.8 &   996 \\
T-REx:P530 \hfill (1)           &            2.8 &             1.8 &            2.0 &    2.3 &       2.8 &                  2.7 &                   2.3 &                  2.8 &   996 \\
T-REx:P530 \hfill (2, LAMA-UHN)           &            2.8 &             1.8 &            2.0 &    2.3 &       2.8 &                  2.7 &                   2.3 &                  2.8 &   996 \\ \midrule
T-REx:P740 \hfill (0, original LAMA)           &            7.6 &            10.5 &           14.7 &    0.0 &      10.4 &                  6.0 &                  13.1 &                 10.4 &   936 \\
T-REx:P740 \hfill (1)           &            5.9 &            10.3 &           13.5 &    0.0 &       9.0 &                  5.2 &                  12.7 &                  9.5 &   910 \\
T-REx:P740 \hfill (2, LAMA-UHN)           &            5.9 &            10.3 &           13.5 &    0.0 &       9.0 &                  5.2 &                  12.7 &                  9.5 &   910 \\ \midrule
T-REx:P937 \hfill (0, original LAMA)           &           29.8 &            33.0 &           38.8 &   40.0 &      32.3 &                 24.9 &                  28.3 &                 34.5 &   954 \\
T-REx:P937 \hfill (1)           &           29.9 &            32.9 &           38.7 &   39.9 &      32.2 &                 24.8 &                  28.2 &                 34.4 &   950 \\
T-REx:P937 \hfill (2, LAMA-UHN)           &           29.9 &            32.9 &           38.7 &   39.9 &      32.2 &                 24.8 &                  28.2 &                 34.4 &   950 \\\midrule
T-REx:P1001 \hfill (0, original LAMA)          &           70.5 &            56.9 &           76.0  &   75.7 &      73.0 &                 73.3 &                  49.5 &                 78.0 &   701 \\
T-REx:P1001 \hfill (1)          &           38.1 &            67.7 &           66.7 &   65.6 &      43.4 &                 40.7 &                  60.3 &                 66.7 &   189 \\
T-REx:P1001 \hfill (2, LAMA-UHN)          &           38.1 &            67.7 &           66.7 &   65.6 &      43.4 &                 40.7 &                  60.3 &                 66.7 &   189 \\ \midrule
T-REx:P1303 \hfill (0, original LAMA)          &            7.6 &            20.3 &           26.6 &    5.3 &       9.1 &                 12.5 &                  29.7 &                 33.2 &   949 \\
T-REx:P1303 \hfill (1)          &            7.6 &            20.3 &           26.6 &    5.3 &       9.1 &                 12.5 &                  29.7 &                 33.2 &   949 \\
T-REx:P1303 \hfill (2, LAMA-UHN)          &            7.6 &            20.3 &           26.6 &    5.3 &       9.1 &                 12.5 &                  29.7 &                 33.2 &   949 \\ \midrule
T-REx:P1376 \hfill (0, original LAMA)          &           73.9 &            41.5 &           62.0 &   71.8 &      75.2 &                 82.1 &                  47.4 &                 70.1 &   234 \\
T-REx:P1376 \hfill (1)          &           74.8 &            42.2 &           62.8 &   73.4 &      75.2 &                 83.5 &                  48.6 &                 72.0 &   218 \\
T-REx:P1376 \hfill (2, LAMA-UHN)          &           74.8 &            42.2 &           62.8 &   73.4 &      75.2 &                 83.5 &                  48.6 &                 72.0 &   218 \\ \midrule
T-REx:P1412 \hfill (0, original LAMA)          &           65.0 &            54.0 &           67.8 &   73.1 &      69.2 &                 63.6 &                  49.3 &                 61.2 &   969 \\
T-REx:P1412 \hfill (1)          &           65.0 &            54.0 &           67.8 &   73.1 &      69.2 &                 63.6 &                  49.3 &                 61.2 &   969 \\
T-REx:P1412 \hfill (2, LAMA-UHN)          &           37.7 &            42.9 &           47.4 &   69.2 &      65.7 &                 51.5 &                  43.5 &                 54.8 &   361 \\ \midrule
Google-RE:date\_of\_birth \hfill (0) &            1.6 &             1.5 &            1.9 &    1.9 &       2.4 &                  1.5 &                   1.5 &                  1.3 &  1825 \\
Google-RE:date\_of\_birth \hfill (1)  &            1.6 &             1.5 &            1.9 &    1.9 &       2.4 &                  1.5 &                   1.5 &                  1.3 &  1825 \\
Google-RE:date\_of\_birth \hfill (2) &            1.6 &             1.5 &            1.9 &    1.9 &       2.4 &                  1.5 &                   1.5 &                  1.3 &  1825 \\ \midrule
Google-RE:place\_of\_birth \hfill (0) &           14.9 &            16.2 &           16.9 &   17.7 &      17.4 &                 16.1 &                  14.8 &                 16.6 &  2937 \\
Google-RE:place\_of\_birth \hfill (1) &           14.9 &            16.2 &           16.8 &   17.7 &      17.4 &                 16.0 &                  14.8 &                 16.6 &  2934 \\
Google-RE:place\_of\_birth \hfill (2) &            5.9 &             9.4 &            8.2 &   10.3 &       9.4 &                  7.2 &                   8.5 &                  7.9 &  2451 \\ \midrule
Google-RE:place\_of\_death \hfill (0) &           13.1 &            12.8 &           14.9 &    6.4 &      13.4 &                 14.0 &                  17.0 &                 14.9 &   766 \\
Google-RE:place\_of\_death \hfill (1) &           13.1 &            12.8 &           14.9 &    6.4 &      13.4 &                 14.0 &                  17.0 &                 14.9 &   766 \\
Google-RE:place\_of\_death \hfill (2) &            6.6 &             7.5 &            7.8 &    2.0 &       7.5 &                  7.6 &                  11.8 &                  8.9 &   655 \\
\bottomrule
\end{tabularx}
\caption{Mean Hits@1 and number of questions per LAMA relation (cont'd). 0: original LAMA dataset, 1: after applying heuristic 1 (string match filter), 2: after applying both heuristics (LAMA-UHN).}
\end{table*}
\end{document}